\documentclass[11pt]{article}
\usepackage{graphicx}
\usepackage{tabularx}
\usepackage{url}
\usepackage{amsthm}
\usepackage{amsmath}
\usepackage{amsfonts}
\usepackage[labelfont=bf]{caption}

\theoremstyle{plain}
\newtheorem{theorem}{Theorem}
\newtheorem{lemma}{Lemma}

\theoremstyle{definition}
\newtheorem{definition}{Definition}

\theoremstyle{remark}
\newtheorem{remark}{Remark}

\newtheorem{exmp}{Example}

\date{}

\begin{document}

\title{Construction of Decision Trees and Acyclic Decision Graphs from Decision Rule Systems}

\author{Kerven Durdymyradov and Mikhail Moshkov \\
Computer, Electrical and Mathematical Sciences \& Engineering Division \\ and Computational Bioscience Research Center\\
King Abdullah University of Science and Technology (KAUST) \\
Thuwal 23955-6900, Saudi Arabia\\ \{kerven.durdymyradov,mikhail.moshkov\}@kaust.edu.sa
}

\maketitle

\begin{abstract}
Decision trees and systems of decision rules are widely
used as classifiers, as a means for knowledge representation, and as
algorithms. They are among the most interpretable models for data analysis.
The study of the relationships between these two models can be seen as an
important task of computer science. Methods for transforming decision trees
into systems of decision rules are simple and well-known. In this paper, we
consider the inverse transformation problem, which is not trivial. We study
the complexity of constructing decision trees and acyclic decision graphs
representing decision trees from decision rule systems, and we discuss the
possibility of not building the entire decision tree, but describing the
computation path in this tree for the given input.
\end{abstract}

{\it Keywords}: decision rule system, decision tree, acyclic decision graph.
\section{Introduction\label{S1}}
In this paper, we consider the problems of transforming systems of decision
rules into decision trees. This paper builds upon our previous work \cite%
{Kerven23}. In that paper, we showed that the minimum depth of a decision
tree derived from the decision rule system can be much less than the number
of different attributes in the rules from the system. In such cases, it is
reasonable to use decision trees.

In the present paper, for some types of decision
rule systems and problems, we prove the existence of polynomial time
algorithms for the construction of decision trees and two types of acyclic
decision graphs representing decision trees. In all other cases, we prove
the absence of such algorithms using the fact that the minimum number of
nodes in decision trees or acyclic decision graphs can grow as a
superpolynomial function depending on the size of decision rule systems. To
avoid difficulties related to the number of nodes in the decision trees, we
discuss also the possibility of not building the entire decision tree, but
describing the computation path in this tree for the given input.

Decision trees \cite%
{AbouEishaACHM19,AlsolamiACM20,BreimanFOS84,Moshkov05,MoshkovZ11,RokachM07}
and decision rule systems \cite%
{BorosHIK97,BorosHIKMM00,ChikalovLLMNSZ13,FurnkranzGL12,MPZ08,MoshkovZ11,Pawlak91,PawlakS07}
are widely used as classifiers, as a means for knowledge representation, and
as algorithms. They are among the most interpretable models for data
analysis \cite{CaoSJ20,GilmoreEH21,Molnar22,SilvaGKJS20}.

The study of the relationships between these two models can be seen as an
important task of computer science. Methods for transforming decision trees
into systems of decision rules are simple and well known \cite%
{Quinlan87,Quinlan93,Quinlan99}. In this paper, we consider the inverse
transformation problem, which is not trivial.

The most known directions of research related to this problem are the
following:

\begin{itemize}
\item Two-stage construction of decision trees. First, decision rules are
built based on the input data, and then decision trees or decision
structures (generalizations of decision trees) are built based on the
constructed rules. Details, including explanations of the benefits of this
approach, can be found in \cite%
{AbdelhalimTN16,AbdelhalimTS09,ImamM93a,ImamM93,ImamM96,KaufmanMPW06,MichalskiI94,MichalskiI97,SzydloSM05}%
.

\item Relations between the depth of deterministic and nondeterministic
decision trees for computing Boolean functions \cite%
{BlumI87,HartmanisH87,Moshkov95,Tardos89}. Note that nondeterministic
decision trees can be interpreted as decision rule systems. The minimum
depth of a nondeterministic decision tree for a Boolean function is equal to
its certificate complexity \cite{BuhrmanW02}.

\item Relations between the depth of deterministic and nondeterministic
decision trees for problems over finite and infinite information systems,
each of which consists of a universe and a set of attributes defined on it
\cite{Moshkov96,Moshkov00,Moshkov03,Moshkov05a,Moshkov20}.
\end{itemize}

This paper continues the development of the so-called syntactic approach to
the study of the considered problem proposed in works \cite%
{Moshkov98,Moshkov01}. This approach assumes that we do not know input data
but only have a system of decision rules that must be transformed into a
decision tree.

Let there be a system of decision rules $S$ of the form $$(a_{i_{1}}=\delta
_{1})\wedge \cdots \wedge (a_{i_{m}}=\delta _{m})\rightarrow \sigma ,$$ where
$a_{i_{1}},\ldots ,a_{i_{m}}$ are attributes, $\delta _{1},\ldots ,\delta
_{m}$ are values of these attributes, and $\sigma $ is a decision. We
describe three problems associated with this system:

\begin{itemize}
\item For a given input (a tuple of values of all attributes included in $S$%
), it is necessary to find all the rules that are realizable for this input
(having a true left-hand side), or show that there are no such rules.

\item For a given input, it is necessary to find all the right-hand sides of
rules that are realizable for this input, or show that there are no such
rules.

\item For a given input, it is necessary to find at least one rule that is
realizable for this input or show that there are no such rules.
\end{itemize}

For each problem, we consider two its variants. The first assumes that in
the input each attribute can have only those values that occur for this
attribute in the system $S$. In the second case, we assume that in the input
any attribute can have any value.

Our goal is to minimize the number of queries for attribute values. For this purpose, decision trees are studied as algorithms for
solving the considered six problems.

 For each of these problems, we investigated in \cite{Kerven23}  unimprovable
upper and lower bounds on the minimum depth of decision trees depending on
three parameters of the decision rule system -- the total number of
different attributes in the rules belonging to the system, the maximum
length of a decision rule, and the maximum number of attribute values.

We proved that, for each problem, there are systems of decision rules for
which the minimum depth of the decision trees that solve the problem is much
less than the total number of attributes in the rule system. For such
systems of decision rules, it is reasonable to use decision trees.

In the present paper, we investigate for each of the considered problems

\begin{itemize}
\item Complexity of constructing decision trees and acyclic decision graphs
representing decision trees.

\item Opportunities not to build the entire decision tree, but to describe
the computation path in this tree for the given input.
\end{itemize}

We prove that in many cases the minimum number of nodes in the decision
trees can grow as a superpolynomial function depending on the size of the
decision rule systems. We show that this issue can be resolved using two
kinds of acyclic decision graphs representing decision trees. However, in
this case, it is necessary to simultaneously minimize the depth and the
number of nodes in acyclic graphs, which is a difficult bi-criteria
optimization problem. We leave this problem for future research and study a
different approach to deal with decision trees: instead of building the
entire decision tree, we model the work of the decision tree for a given
tuple of attribute values using a polynomial time algorithm.

In this paper, we
repeat the main definitions from \cite{Kerven23} and give some lemmas from
\cite{Kerven23} without proofs. In Remarks \ref{R3} and \ref{R4} (see Section \ref{S7}), we
mention some results that were published in \cite{Moshkov01} without proofs and discussed in details in the present paper.

This paper consists of eight sections. Section \ref{S2} discusses the main
definitions and notation. Sections \ref{S3}-\ref{S6} consider the problems
of constructing decision trees and acyclic decision graphs representing decision 
trees. Section \ref{S7} discusses the possibility to construct not the
entire decision tree, but the computation path in this tree for the given
input. Section \ref{S8} contains a short conclusion.

\section{Main Definitions and Notation\label{S2}}

In this section, we discuss the main definitions and notation related to
decision rule systems and decision trees. In fact, we repeat the definitions and notation from \cite{Kerven23}, but consider new examples.

\subsection{Decision Rule Systems\label{S2.1}}

Let $\omega =\{0,1,2,\ldots \}$ and $A=\{a_{i}:i\in \omega \}$. Elements of
the set $A$ will be called \emph{attributes}. 

\begin{definition}
A \emph{decision rule} is an expression of the form
\begin{equation*}
(a_{i_{1}}=\delta _{1})\wedge \cdots \wedge (a_{i_{m}}=\delta
_{m})\rightarrow \sigma ,
\end{equation*}%
where $m\in \omega $, $a_{i_{1}},\ldots ,a_{i_{m}}$ are pairwise different
attributes from $A$ and $\delta _{1},\ldots ,\delta _{m},\sigma \in \omega $. 
\end{definition}

We denote this decision rule by $r$. The expression $(a_{i_{1}}=\delta
_{1})\wedge \cdots \wedge (a_{i_{m}}=\delta _{m})$ will be called the \emph{%
left-hand side}, and the number $\sigma $ will be called the \emph{%
right-hand side} of the rule $r$. The number $m$ will be called the \emph{%
length }of the decision rule $r$. Denote $A(r)=\{a_{i_{1}},\ldots
,a_{i_{m}}\}$ and $K(r)=\{a_{i_{1}}=\delta _{1},\ldots ,a_{i_{m}}=\delta
_{m}\}$. If $m=0$, then $A(r)=K(r)=\emptyset $.

\begin{definition}
Two decision rules $r_1$ and $r_2$ are \emph{equal} if $K(r_1)=K(r_2)$ and the right-hand sides of the rules $r_1$ and $r_2$ are equal.
\end{definition}

\begin{definition}
A \emph{system of decision rules} $S$ is a finite nonempty set of decision
rules. 
\end{definition}

Denote $A(S)=\bigcup_{r\in S}A(r)$, $n(S)=\left\vert A(S)\right\vert $%
, $D(S)$ the set of the right-hand sides of decision rules from $S$, and $%
d(S)$ the maximum length of a decision rule from $S$. Let $n(S)>0$. For $a_{i}\in A(S)$,
let $V_{S}(a_{i})=\{\delta :a_{i}=\delta \in \bigcup_{r\in S}K(r)\}$ and $%
EV_{S}(a_{i})=V_{S}(a_{i})\cup \{\ast \}$, where the symbol $\ast $ is
interpreted as a number that does not belong to the set $V_{S}(a_{i})$.
Denote $k(S)=\max \{\left\vert V_{S}(a_{i})\right\vert :a_{i}\in A(S)\}$. If $n(S)=0$, then $k(S)=0$. We
denote by $\Sigma $ the set of systems of decision rules.

\begin{exmp}
Let us consider a decision rule system $S=\{(a_{1}=0)\rightarrow 1, (a_{1}=1)\wedge (a_{2}=0)\rightarrow 2, (a_{1}=2)\wedge (a_{3}=0)\wedge (a_{4}=0)\rightarrow 3\}$. Then $A(r_3)=\{a_{1},a_{3},a_{4}\}$, $K(r_3)=\{a_{1}=2,a_{3}=0,a_{4}=0\}$, where $r_3$ denotes the third rule from $S$. $A(S)=\bigcup_{r\in S}A(r) = \{a_{1},a_{2},a_{3},a_{4}\}$, $n(S)=\left\vert A(S)\right\vert = 4$, $D(S) = \{1,2,3\}$, $d(S) = 3$, $V_{S}(a_{1})=\{\delta :a_{1}=\delta \in \bigcup_{r\in S}K(r)\} = \{0, 1, 2\}$, $EV_{S}(a_{1})=V_{S}(a_{1})\cup \{\ast \} = \{0, 1, 2, \ast\}$ and $k(S)=\max \{\left\vert V_{S}(a_{i})\right\vert :a_{i}\in A(S)\} = \left\vert V_{S}(a_{1})\right\vert = 3$.
\end{exmp}

Let $S\in \Sigma $, $n(S)>0$, and $A(S)=\{a_{j_{1}},\ldots ,a_{j_{n}}\}$, where $%
j_{1}<\cdots <j_{n}$. Denote $V(S)=V_{S}(a_{j_{1}})\times \cdots \times
V_{S}(a_{j_{n}})$ and $EV(S)=EV_{S}(a_{j_{1}})\times \cdots \times
EV_{S}(a_{j_{n}})$. For $\bar{\delta}=(\delta _{1},\ldots ,\delta _{n})\in
EV(S)$, denote $K(S,\bar{\delta})=\{a_{j_{1}}=\delta _{1},\ldots
,a_{j_{n}}=\delta _{n}\}$. 

\begin{definition}
We will say that a decision rule $r$ from $S$ is \emph{%
realizable} for a tuple $\bar{\delta} \in EV(S)$ if $K(r)\subseteq K(S,\bar{\delta})$. 
\end{definition}

It is clear that any rule with an empty left-hand side is realizable for the tuple $\bar{\delta}$.

\begin{exmp}
Let us consider a decision rule system $S  = \{r_1: (a_{1}=0)\wedge (a_{2}=1)\rightarrow 1$, $r_2: (a_{1}=1)\wedge (a_{3}=1)\rightarrow 2\}$ and a tuple $\bar{\delta} = (1,1,1) \in EV(S)$. Then the decision rule $r_2$ from $S$ is realizable for the tuple $\bar{\delta}$, but $r_1$ is not.
\end{exmp}

Let $V\in \{V(S),EV(S)\}$. We now define three problems related to the rule
system $S$.

\begin{definition}
Problem \emph{All Rules} for the pair $(S,V)$: for a given tuple $\bar{\delta%
}\in V$, it is required to find the set of rules from $S$ that are
realizable for the tuple $\bar{\delta}$.
\end{definition}

\begin{definition}
Problem \emph{All Decisions} for the pair $(S,V)$: for a given tuple $\bar{%
\delta}\in V$, it is required to find a set $Z$ of decision rules from $S$
satisfying the following conditions:

\begin{itemize}
\item All decision rules from $Z$ are realizable for the tuple $\bar{\delta}$%
.

\item For any $\sigma \in D(S)\setminus D(Z)$, any decision rule from $S$
with the right-hand side equal to $\sigma $ is not realizable for the tuple $%
\bar{\delta}$.
\end{itemize}
\end{definition}

\begin{definition}
Problem \emph{Some Rules} for the pair $(S,V)$: for a given tuple $\bar{%
\delta}\in V$, it is required to find a set $Z$ of decision rules from $S$
satisfying the following conditions:

\begin{itemize}
\item All decision rules from $Z$ are realizable for the tuple $\bar{\delta}$%
.

\item If $Z=\emptyset $, then any decision rule from $S$ is not realizable
for the tuple $\bar{\delta}$.
\end{itemize}
\end{definition}

Denote $AR(S)$ and $EAR(S)$ the problems All Rules for pairs $(S,V(S))$ and $%
(S,EV(S))$, respectively. Denote $AD(S)$ and $EAD(S)$ the problems All
Decisions for pairs $(S,V(S))$ and $(S,EV(S))$, respectively. Denote $SR(S)$
and $ESR(S)$ the problems Some Rules for pairs $(S,V(S))$ and $(S,EV(S))$,
respectively.

\begin{exmp}
Let a decision rule system $S=\{(a_{1}=0)\rightarrow 1, (a_{1}=0)\wedge (a_{2}=1)\rightarrow 1, (a_{1}=0)\wedge (a_{2}=1)\wedge (a_{3}=2)\rightarrow 2\}$ and a tuple $\bar{\delta}=(0,1,2) \in V(S)$ are given. Then $\{(a_{1}=0)\rightarrow 1, (a_{1}=0)\wedge (a_{2}=1)\rightarrow 1, (a_{1}=0)\wedge (a_{2}=1)\wedge (a_{3}=2)\rightarrow 2\}$ is the solution for the problem $AR(S)$ and the tuple $\bar{\delta}$, $\{(a_{1}=0)\rightarrow 1, (a_{1}=0)\wedge (a_{2}=1)\wedge (a_{3}=2)\rightarrow 2\}$ is a solution for the problem $AD(S)$ and the tuple $\bar{\delta}$, and $\{(a_{1}=0)\rightarrow 1\}$ is a solution for the problem $SR(S)$ and the tuple $\bar{\delta}$.
\end{exmp}

In the special case, when $n(S)=0$,  all rules from $S$ have an empty left-hand side. In this case, it is natural to consider (i) the set $S$ as the solution to the problems $AR(S)$ and $EAR(S)$, (ii) any subset $Z$ of the set $S$ with $D(Z)=D(S)$ as a solution to the problems $AD(S)$ and $EAD(S)$, and (iii) any nonempty subset $Z$ of the set $S$ as a solution to the problems $SR(S)$ and $ESR(S)$.

Let $S\in \Sigma $, where $\Sigma $ is the set of decision rule systems. We
denote by $R_{SR}(S)$ a subsystem of the system $S$ that consists of all
rules $r\in S$ satisfying the following condition: there is no rule $%
r^{\prime }\in S$ such that $K(r^{\prime })\subset K(r)$. 

\begin{definition}
The system $S$ will be called $SR$-\emph{reduced} if $R_{SR}(S)=S$.
\end{definition}

Denote by $\Sigma _{SR}$ the set of $SR$-reduced systems of decision rules.

For $S\in \Sigma $, we denote by $R_{AD}(S)$ a subsystem of the system $S$ that
consists of all rules $r\in S$ satisfying the following condition: there is
no a rule $r^{\prime }\in S$ such that $K(r^{\prime })\subset K(r)$ and the
right-hand sides of the rules $r$ and $r^{\prime }$ coincide. 

\begin{definition}
The system $S$ will be called $AD$-\emph{reduced} if $R_{AD}(S)=S$.
\end{definition}

Denote by $\Sigma _{AD}$ the set of $AD$-reduced systems of decision rules.

\begin{exmp}
Let us consider a decision rule system $S=\{(a_{1}=0)\wedge (a_{2}=1)\wedge (a_{3}=2)\rightarrow 1, (a_{1}=0)\wedge (a_{2}=1)\rightarrow 2, (a_{1}=0)\rightarrow 2\}$. For this system, $R_{AD}(S)=\{(a_{1}=0)\wedge (a_{2}=1)\wedge (a_{3}=2)\rightarrow 1, (a_{1}=0)\rightarrow 2\}$ and $R_{SR}(S)=\{(a_{1}=0)\rightarrow 2\}$.
\end{exmp}

\subsection{Decision Trees\label{S2.2}}

A \emph{finite directed tree with root} is a finite directed tree in which
only one node has no entering edges. This node is called the \emph{root}.
The nodes without leaving edges are called \emph{terminal} nodes. The nodes
that are not terminal will be called \emph{working} nodes. A \emph{complete
path} in a finite directed tree with root is a sequence $\xi
=v_{1},d_{1},\ldots ,v_{m},d_{m},v_{m+1}$ of nodes and edges of this tree in
which $v_{1}$ is the root, $v_{m+1}$ is a terminal node and, for $i=1,\ldots
,m$, the edge $d_{i}$ leaves the node $v_{i}$ and enters the node $v_{i+1}$.

We will consider two types of decision trees: o-decision trees (ordinary
decision trees, o-trees in short) and e-decision trees (extended decision
trees, e-trees in short). 

\begin{definition}
A \emph{decision tree over a decision rule system}
$S$ is a labeled finite directed tree with root $\Gamma $ satisfying the
following conditions:

\begin{itemize}
\item Each working node of the tree $\Gamma $ is labeled with an attribute
from the set $A(S)$.

\item Let a working node $v$ of the tree $\Gamma $ be labeled with an
attribute $a_{i}$. If $\Gamma $ is an o-tree, then exactly $\left\vert
V_{S}(a_{i})\right\vert $ edges leave the node $v$ and these edges are
labeled with pairwise different elements from the set $V_{S}(a_{i})$. If $%
\Gamma $ is an e-tree, then exactly $\left\vert EV_{S}(a_{i})\right\vert $
edges leave the node $v$ and these edges are labeled with pairwise different
elements from the set $EV_{S}(a_{i})$.

\item Each terminal node of the tree $\Gamma$ is labeled with a subset of
the set $S$.
\end{itemize}
\end{definition}

Let $\Gamma $ be a decision tree over the decision rule system $S$. We
denote by $CP(\Gamma )$ the set of complete paths in the tree $\Gamma $. Let
$\xi =v_{1},d_{1},\ldots ,v_{m},d_{m},v_{m+1}$ be a complete path in $\Gamma
$. We correspond to this path a set of attributes $A(\xi )$ and an equation
system $K(\xi )$. If $m=0$ and $\xi =v_{1}$, then $A(\xi )=\emptyset $ and $K(\xi )=\emptyset
$. Let $m>0$ and, for $j=1,\ldots ,m$, the node $v_{j}$ be labeled with the
attribute $a_{i_{j}}$ and the edge $d_{j}$ be labeled with the element $%
\delta _{j}\in \omega \cup \{\ast \}$. Then $A(\xi )=\{a_{i_{1}},\ldots
,a_{i_{m}}\}$ and $K(\xi )=\{a_{i_{1}}=\delta _{1},\ldots ,a_{i_{m}}=\delta
_{m}\}$. We denote by $\tau (\xi )$ the set of decision rules attached to
the node $v_{m+1}$.

\begin{exmp}
Let us consider a decision rule system $S=\{(a_{1}=0)\wedge (a_{2}=0)\rightarrow 1, (a_{1}=0)\wedge (a_{2}=1)\rightarrow 2, (a_{1}=1)\wedge (a_{3}=0)\rightarrow 3, (a_{1}=1)\rightarrow 4\}$. Then o-tree $\Gamma_1$ and e-tree $\Gamma_2$ over the decision rule system $S$ are given in Fig. \ref{fig:fig1}, where $r_1$, $r_2$, $r_3$ and $r_4$ are the first, second, third and last decision rules in $S$, respectively.

\begin{figure}[h!]
\renewcommand{\figurename}{\textbf{Fig.}}
    \centering
    \includegraphics[width=0.7\textwidth]{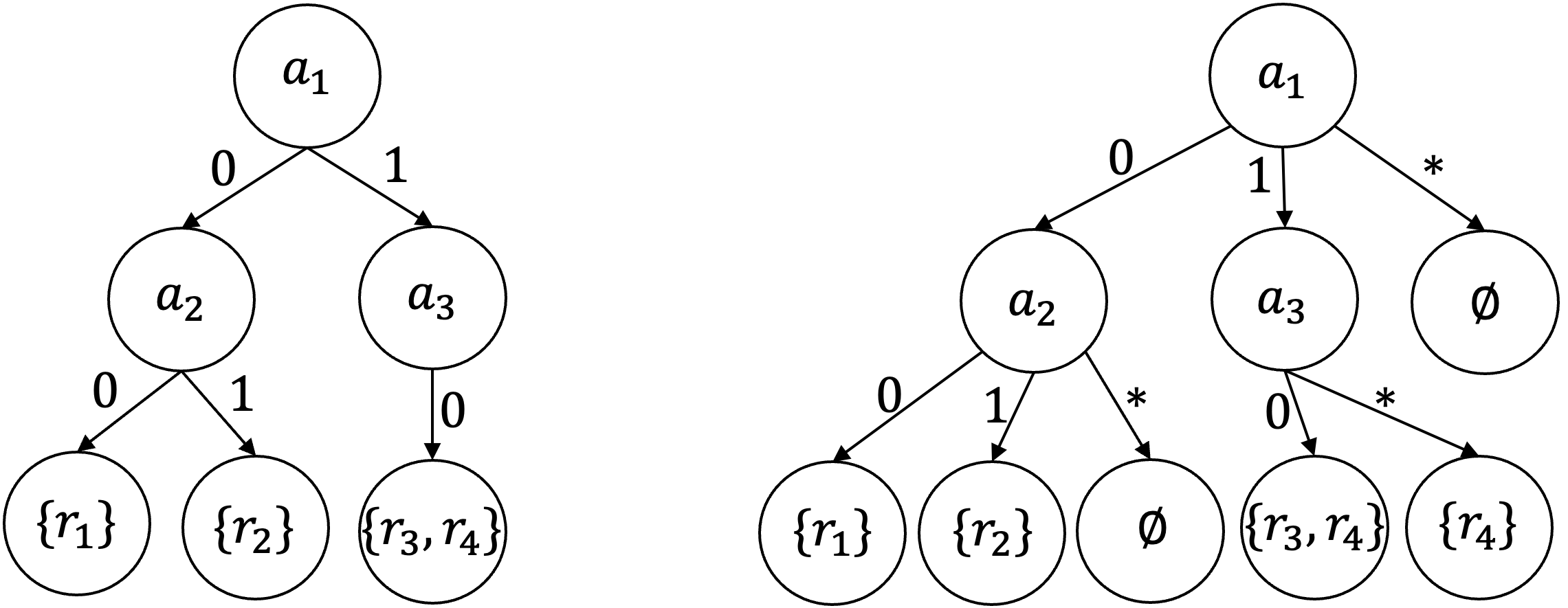}
    \caption{o-decision tree $\Gamma_1$ and e-decision tree $\Gamma_2$ over the decision rule system $S$}
    \label{fig:fig1}
\end{figure}
Let $\xi$ be a complete path in the o-tree $\Gamma_1$, which is finished in the terminal node labeled with the set of rules $\{r_3, r_4\}$. Then $A(\xi )=\{a_{1}, a_{3}\}$,  $K(\xi )=\{a_{1}=1, a_{3}=0\}$ and  $\tau (\xi ) = \{r_{3}, r_{4}\}$.
\end{exmp}

\begin{definition}
A system of equations $\{a_{i_{1}}=\delta _{1},\ldots ,a_{i_{m}}=\delta
_{m}\}$, where $a_{i_{1}},\ldots ,a_{i_{m}}\in A$ and $\delta _{1},\ldots
,\delta _{m}\in \omega \cup \{\ast \}$, will be called \emph{inconsistent}
if there exist $l,k\in \{1,\ldots ,m\}$ such that $l\neq k$, $i_{l}=i_{k}$,
and $\delta _{l}\neq \delta _{k}$. If the system of equations is not
inconsistent, then it will be called \emph{consistent}.
\end{definition}

Let $S$ be a decision rule system and $\Gamma $ be a decision tree over $S$.

\begin{definition}
We will say that $\Gamma $ \emph{solves} the problem $AR(S)$ (the problem $EAR(S)$, respectively) if $\Gamma $ is an o-tree (an e-tree, respectively)
and any path $\xi \in CP(\Gamma )$ with consistent system of equations $%
K(\xi )$ satisfies the following conditions:

\begin{itemize}
\item For any decision rule $r\in \tau (\xi )$, the relation $K(r)\subseteq
K(\xi )$ holds.

\item For any decision rule $r\in S\setminus \tau (\xi )$, the system of
equations $K(r)\cup K(\xi )$ is inconsistent.
\end{itemize}
\end{definition}

\begin{exmp}
Let $S$ be a decision rule system from Example 4. Then the decision trees $\Gamma_1$ and $\Gamma_2$ depicted in Fig. \ref{fig:fig1} solve the problems $AR(S)$ and $EAR(S)$, respectively.
\end{exmp}

\begin{definition}
We will say that $\Gamma $ \emph{solves} the problem $AD(S)$ (the problem $%
EAD(S)$, respectively) if $\Gamma $ is an o-tree (an e-tree, respectively)
and any path $\xi \in CP(\Gamma )$ with consistent system of equations $%
K(\xi )$ satisfies the following conditions:

\begin{itemize}
\item For any decision rule $r\in \tau (\xi )$, the relation $K(r)\subseteq
K(\xi )$ holds.

\item If $r\in S\setminus \tau (\xi )$ and the right-hand side of $r$ does
not belong to the set $D(\tau (\xi ))$, then the system of equations $%
K(r)\cup K(\xi )$ is inconsistent.
\end{itemize}
\end{definition}

\begin{definition}
We will say that $\Gamma $ \emph{solves} the problem $SR(S)$ (the problem $%
ESR(S)$, respectively) if $\Gamma $ is an o-tree (an e-tree, respectively)
and any path $\xi \in CP(\Gamma )$ with consistent system of equations $%
K(\xi )$ satisfies the following conditions:

\begin{itemize}
\item For any decision rule $r\in \tau (\xi )$, the relation $K(r)\subseteq
K(\xi )$ holds.

\item If $\tau (\xi )=\emptyset $, then, for any decision rule $r\in S$, the
system of equations $K(r)\cup K(\xi )$ is inconsistent.
\end{itemize}
\end{definition}

\begin{exmp}
Let us consider a decision rule system $S=\{(a_{1}=0)\wedge (a_{2}=0)\rightarrow 1, (a_{1}=0)\wedge (a_{2}=1)\rightarrow 2, (a_{1}=1)\wedge (a_{3}=0)\rightarrow 3, (a_{1}=1)\rightarrow 3, (a_{1}=1)\rightarrow 4\}$ and decision trees $\Gamma_1$, $\Gamma_2$, and $\Gamma_3$ depicted in Fig. \ref{fig:fig2}. Then the decision tree $\Gamma_1$ solves the problem $AR(S)$, $\Gamma_2$ solves $AD(S)$, and $\Gamma_3$ solves $SR(S)$, where $r_1$, $r_2$, $r_3$, $r_4$ and $r_5$ are the first, second, third, fourth and fifth decision rules in $S$, respectively.

\begin{figure}[h!]
\renewcommand{\figurename}{\textbf{Fig.}}
    \centering
    \includegraphics[width=0.85\textwidth]{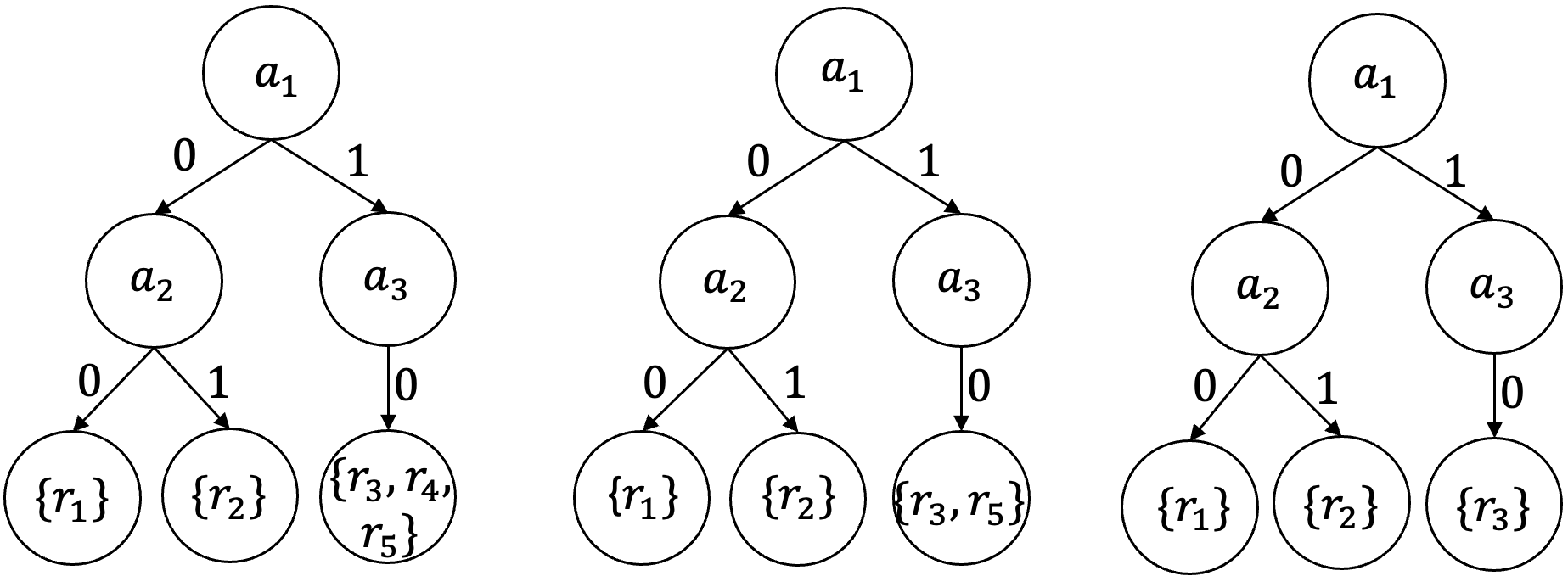}
    \caption{Decision trees $\Gamma_1$, $\Gamma_2$, and $\Gamma_3$}
    \label{fig:fig2}
\end{figure}

\end{exmp}

For any complete path $\xi \in CP(\Gamma )$, we denote by $h(\xi )$ the
number of working nodes in $\xi $. The value $h(\Gamma )=\max \{h(\xi ):\xi
\in CP(\Gamma )\}$ is called the \emph{depth} of the decision tree $\Gamma $.

Let $S$ be a decision rule system and $C\in \{AR,EAR,AD,EAD,SR,ESR\}$. We
denote by $h_{C}(S)$ the minimum depth of a decision tree over $S$,
which solves the problem $C(S)$.

Let $n(S)=0$. If $C \in \{AR,EAR\}$, then there is only one decision tree solving the problem $C(S)$. This tree consists of one node labeled with the set of rules $S$. If $C \in \{AD,EAD\}$, then the set of decision trees solving the problem $C(S)$ coincides with the set of trees each of which consists of one node labeled with a subset $Z$ of the set $S$ with $D(Z)=D(S)$. If $C \in \{SR,ESR\}$, then the set of decision trees solving the problem $C(S)$ coincides with the set of trees each of which consists of one node labeled with a nonempty subset $Z$ of the set $S$. Therefore if $n(S)=0$, then $h_{C}(S)=0$ for any $C\in \{AR,EAR,AD,EAD,SR,ESR\}$.

\section{Auxiliary Statements\label{S3}}
In this section, we will first give some statements from \cite{Kerven23} and then we will prove some new ones.

Let $S$ be a decision rule system and $\Gamma $ be an e-decision tree over $%
S $. We denote by $o(\Gamma )$ an o-tree over $S$, which is obtained from
the tree $\Gamma $ by the removal of all nodes $v$ such that the path from
the root to the node $v$ in $\Gamma $ contains an edge labeled with $\ast $.
Together with a node $v$, we remove all edges entering or leaving $v$.

\begin{lemma}
\label{L1} (Lemma 1 \cite{Kerven23}) Let $S$ be a decision rule system, $C\in \{AR,AD,SR\}$, and $%
\Gamma $ be an e-decision tree over $S$ solving the problem $EC(S)$. Then the
decision tree $o(\Gamma )$ solves the problem $C(S)$.
\end{lemma}

\begin{lemma}
\label{L2} (Lemma 2 \cite{Kerven23}) Let $S$ be a decision rule system and $\Gamma $ be a decision
tree over $S$. Then

(a) If the tree $\Gamma $ solves the problem $AR(S)$ ($EAR(S)$,
respectively), then the tree $\Gamma $ solves the problem $AD(S)$ ($EAD(S)$,
respectively).

(b) If the tree $\Gamma $ solves the problem $AD(S)$ ($EAD(S)$,
respectively), then the tree $\Gamma $ solves the problem $SR(S)$ ($ESR(S)$,
respectively).
\end{lemma}

\begin{lemma}
\label{L3} (Lemma 3 \cite{Kerven23}) Let $S$ be a decision rule system. Then the following
inequalities hold:

\begin{equation*}
\begin{array}{ccccccc}
h_{ESR}(S) & \leq & h_{EAD}(S) & \leq & h_{EAR}(S) & \leq & n(S) \\
\mathrel{\rotatebox{90}{$\le$}} &  & \mathrel{\rotatebox{90}{$\le$}} &  & %
\mathrel{\rotatebox{90}{$\le$}} &  &  \\
h_{SR}(S) & \leq & h_{AD}(S) & \leq & h_{AR}(S) &  &
\end{array}%
\end{equation*}
\end{lemma}

Let $S$ be a decision rule system
and $\alpha =\{a_{i_{1}}=\delta _{1},\ldots ,a_{i_{m}}=\delta _{m}\}$ be a
consistent equation system such that $a_{i_{1}},\ldots ,a_{i_{m}}\in A$ and $%
\delta _{1},\ldots ,\delta _{m}\in \omega \cup \{\ast \}$. We now define a
decision rule system $S_{\alpha }$. Let $r$ be a decision rule for which the
equation system $K(r)\cup \alpha $ is consistent. We denote by $r_{\alpha }$
the decision rule obtained from $r$ by the removal from the left-hand side
of $r$ all equations that belong to $\alpha $. Then $S_{\alpha }$ is the set
of decision rules $r_{\alpha }$ such that $r\in S$ and the equation system $%
K(r)\cup \alpha $ is consistent.

\begin{lemma}
\label{L4} (Lemma 6 \cite{Kerven23}) Let $S$ be a decision rule system with $n(S)>0$, $C\in \{EAR,AR,\\
EAD,AD,ESR,$ $SR\}$%
, $\alpha =\{a_{i_{1}}=\delta _{1},\ldots ,a_{i_{m}}=\delta _{m}\}$ be a
consistent equation system such that $a_{i_{1}},\ldots ,a_{i_{m}}\in A(S)$
and, for $j=1,\ldots ,m$,  $\delta _{j}\in
EV_{S}(a_{i_{j}})$ if $C\in \{EAR,EAD,ESR\}$ and $\delta _{j}\in
V_{S}(a_{i_{j}})$ if $C\in \{AR,AD,SR\}$. Then $h_{C}(S)\geq h_{C}(S_{\alpha })$.
\end{lemma}

We correspond to a decision rule system $S$ a hypergraph $G(S)$ with the set of nodes $%
A(S)$ and the set of edges $\{A(r):r\in S\}$. A \emph{node cover} of the
hypergraph $G(S)$ is a subset $B$ of the set of nodes $A(S)$ such that $%
A(r)\cap B\neq \emptyset $ for any rule $r\in S$ such that $A(r)\neq
\emptyset $. If $A(S)=\emptyset $, then the empty set is the only node cover of the
hypergraph $G(S)$. Denote by $\beta (S)$ the minimum cardinality of a node
cover of the hypergraph $G(S)$.

\begin{exmp}
Let us consider a decision rule system $S=\{(a_{1}=0)\wedge (a_{2}=1)\rightarrow 1, (a_{1}=0)\wedge (a_{3}=1)\rightarrow 2, (a_{4}=0)\rightarrow 3\}$. One can show, that the set $\{a_{1}, a_{4}\}$ is a node cover of the hypergraph $G(S)$ and $\beta (S) = 2$.
\end{exmp}

We define a subsystem $I_{SR}(S)$ of the system $S$ in the following way. If
$S$ does not contain rules of the length $0$, then $I_{SR}(S)=S$. Otherwise,
$I_{SR}(S)$ consists of all rules from $S$ of the length $0$.

\begin{remark}
\label{R1}Note that $I_{SR}(S)\neq S$ if and only if $S$ contains both a
rule of the length $0$ and a rule of the length greater than $0$.
\end{remark}

We now define a subsystem $I_{AD}(S)$ of the system $S$. Denote by $D_{0}(S)$
the set of the right-hand sides of decision rules from $S$, which length is
equal to $0$. Then the subsystem $I_{AD}(S)$ consists of all rules from $S$
of the length $0$ and all rules from $S$ for which the right-hand sides do
not belong to $D_{0}(S)$.

\begin{remark}
\label{R2}Note that $I_{AD}(S)\neq S$ if and only if $S$ contains both a
rule of the length $0$ and a rule of the length greater than $0$ with the
same right-hand sides.
\end{remark}

\begin{exmp}
Let us consider a decision rule system $S=\{(a_{1}=0)\rightarrow 1, (a_{2}=0)\rightarrow 2, \rightarrow 2\}$. For this system, $I_{SR}(S)=\{\rightarrow 2\}$ and $I_{AD}(S)=\{(a_{1}=0)\rightarrow 1, \rightarrow 2\}$.
\end{exmp}

Let $S$ be a decision rule system. This system will be called \emph{%
incomplete} if there exists a tuple $\bar{\delta}\in V(S)$ such that the
equation system $K(r)\cup K(S,\bar{\delta})$ is inconsistent for any
decision rule $r\in S$. Otherwise, the system $S$ will be called \emph{complete}. If $n(S)=0$, then the system $S$ will be considered as complete.

\begin{lemma}
\label{L5} (Lemma 7 \cite{Kerven23}) Let $S$ be a decision rule system. Then

(a) If $C\in \{EAR,AR\}$, then $h_{C}(S)\geq \beta (S)$.

(b) If $C\in \{AD,SR\}$, then $h_{EC}(S)\geq \beta (I_{C}(S))$.

(c) If $C\in \{AD,SR\}$ and the system $S$ is incomplete, then $h_{C}(S)\geq
\beta (S)$.
\end{lemma}

\begin{lemma}
\label{L6} (Lemma 8 \cite{Kerven23}) Let $S$ be a decision rule system. Then

(a) $h_{EAR}(S)\geq h_{AR}(S)\geq d(S)$.

(b) If $S$ is an $SR$-reduced system, then $h_{ESR}(S)\geq d(S)$.

(c) If $S$ is an $AD$-reduced system, then $h_{EAD}(S)\geq d(S)$.
\end{lemma}

\begin{lemma}
\label{L7} (Lemma 9 \cite{Kerven23}) Let $S$ be a decision rule system. Then $%
h_{ESR}(S)=h_{ESR}(R_{SR}(S))$ and $h_{EAD}(S)=h_{EAD}(R_{AD}(S))$.
\end{lemma}

Let $C\in \{AR,EAR,AD,EAD,SR,ESR\}$, $S$ be a decision rule system, and $%
\Gamma $ be a decision tree over $S$. We denote by $L(\Gamma )$ the number
of nodes in the tree $\Gamma $ and by $T(\Gamma )$ we denote the number of terminal
nodes in $\Gamma $ that are labeled with pairwise different sets of decision
rules. Let $L_{C}(S)=\min L(\Gamma )$ and $T_{C}(S)=\min T(\Gamma )$, where
the minimum is taken over all decision trees $\Gamma $ over $S$ that solve
the problem $C(S)$. It is clear that $T_{C}(S)\leq L_{C}(S)$.

\begin{lemma}
\label{L8}Let $S$ be a decision rule system. Then the following
inequalities hold: \smallskip

$%
\begin{array}{ccccc}
L_{ESR}(S) & \leq & L_{EAD}(S) & \leq & L_{EAR}(S) \\
\mathrel{\rotatebox{90}{$\le$}} &  & \mathrel{\rotatebox{90}{$\le$}} &  & %
\mathrel{\rotatebox{90}{$\le$}} \\
L_{SR}(S) & \leq & L_{AD}(S) & \leq & L_{AR}(S),%
\end{array}%
$
\smallskip
$%
\begin{array}{ccccc}
T_{ESR}(S) & \leq & T_{EAD}(S) & \leq & T_{EAR}(S) \\
\mathrel{\rotatebox{90}{$\le$}} &  & \mathrel{\rotatebox{90}{$\le$}} &  & %
\mathrel{\rotatebox{90}{$\le$}} \\
T_{SR}(S) & \leq & T_{AD}(S) & \leq & T_{AR}(S).%
\end{array}%
$
\end{lemma}

\begin{proof}
It is clear that the considered inequalities hold if $n(S)=0$. Let $n(S)>0$.

Let $\Gamma $ be an e-decision tree over $S$. It is clear that $L(o(\Gamma
))\leq L(\Gamma )$ and $T(o(\Gamma ))\leq T(\Gamma )$. Using these
inequalities and Lemma \ref{L1}, we obtain that $L_{SR}(S)\leq L_{ESR}(S)$, $%
L_{AD}(S)\leq L_{EAD}(S)$, $L_{AR}(S)\leq L_{EAR}(S)$, $T_{SR}(S)\leq
T_{ESR}(S)$, $T_{AD}(S)\leq T_{EAD}(S)$, and $T_{AR}(S)\leq T_{EAR}(S)$.

Using Lemma \ref{L2}, we obtain that $L_{ESR}(S)\leq L_{EAD}(S)\leq
L_{EAR}(S)$, $L_{SR}(S)\leq L_{AD}(S)\leq L_{AR}(S)$, $T_{ESR}(S)\leq
T_{EAD}(S)\leq T_{EAR}(S)$, and $T_{SR}(S)\leq T_{AD}(S)\leq T_{AR}(S)$.
\end{proof}

A system of decision rules $S$ can be represented by a word over the alphabet $$
\{(,),a,=,\wedge ,\rightarrow ,0,1,;\}$$ in which numbers from $\omega $
(attribute indexes, attribute values, and right-hand sides of decision
rules) are in binary representation (are represented by words over the
alphabet $\{0,1\}$) and the symbol \textquotedblleft ;" is used to separate
two rules. The length of this word will be called the \emph{size} of the
decision rule system $S$ and will be denoted $size(S)$.

\begin{definition}
A decision rule system $S$ will be called \emph{reduced} if it satisfies the
following conditions:

\begin{itemize}
\item If $n(s)=n$, then $A(S)\subseteq \{a_{0},\ldots ,a_{n}\}$.

\item If $\left\vert
D(S)\right\vert =t$, then $D(S)\subseteq\{0,\ldots ,t\}$.

\item If $k(S)=k$, then $V_{S}(a_{i})\subseteq \{0,\ldots ,k\}$ for any $a_{i}\in A(S)$.

\item $d(S) \ge 1$.
\end{itemize}
\end{definition}

Let $S$ be a reduced system of decision rules. It is clear that $n(S)\leq d(S)\left\vert
S\right\vert $, $k(S)\leq \left\vert S\right\vert $, and $\left\vert
D(S)\right\vert \leq \left\vert S\right\vert $. Therefore the maximum
number from $\omega $ in the system $S$ is at most $d(S)\left\vert
S\right\vert $. The length of binary representation of such a number is at
most $\log _{2}(d(S)\left\vert S\right\vert )+1$. The length of each
rule from $S$ is at most $d(S)$. One can show that the length of word representing
each rule (including the sign ``;" after it) is at most $10d(S) (\log _{2}(d(S)\left\vert
S\right\vert )+1)$ and
\begin{equation}
size(S) \le10\left\vert S\right\vert
d(S)(\log _{2}(d(S)\left\vert S\right\vert )+1). \label{E13}
\end{equation}
\begin{lemma}
\label{L9} Let $n,k,d\in \omega \setminus \{0\}$, $d\geq 2$, and $k\geq 2$.
Then there exists a reduced decision rule system $S$ such that $n(S)=2n+d$, $k(S)=k$%
, $d(S)=d$, $\left\vert S\right\vert =2n+k-1$, and $L_{SR}(S)\geq 2^{n}$.
\end{lemma}

\begin{proof}
Let us consider a system of decision rules $S=S_{1}\cup S_{2}\cup S_{3}$,
where $S_{1}=\{(a_{2i-1}=0)\wedge (a_{2i}=0)\rightarrow 0,(a_{2i-1}=1)\wedge
(a_{2i}=1)\rightarrow 0:i=1,\ldots ,n\}$, $S_{2}=\{(a_{1}=2)\wedge
(a_{2}=2)\rightarrow 0,\ldots ,(a_{1}=k-1)\wedge (a_{2}=k-1)\rightarrow 0\}$%
, and $S_{3}=\{(a_{2n+1}=0)\wedge \cdots \wedge (a_{2n+d}=0)\rightarrow 0\}$%
. If $k=2$, then $S_{2}=\emptyset $. It is clear that $S$ is reduced, $n(S)=2n+d$, $k(S)=k$,
$d(S)=d$, and $\left\vert S\right\vert =2n+k-1$. Denote by $\Delta $ the set
of tuples $(\delta _{1},\ldots ,\delta _{2n+d})\in \{0,1\}^{2n+d}$ such that
$\delta _{2i-1}+\delta _{2i}=1$ for $i=1,\ldots ,n$ and $\delta
_{2n+1}=\cdots =\delta _{2n+d}=1$. It is clear that $\left\vert \Delta
\right\vert =2^{n}$ and, for any $\bar{\delta}\in \Delta $, there is no a
rule from $S$ that is realizable for the tuple $\bar{\delta}$.

Let $\Gamma $
be a decision tree over $S$, which solves the problem $SR(S)$ and for which $%
L(\Gamma )=L_{SR}(S)$. Let $\bar{\delta}=(\delta _{1},\ldots ,\delta
_{2n+d})\in \Delta $. It is clear that there exists a complete path $\xi $ in $%
\Gamma $ such that $K(\xi )\subseteq K(S,\bar{\delta})$. Evidently, the
terminal node of this path is labeled with the empty set. Let us show that $%
\{a_{1}=\delta _{1},\ldots ,a_{2n}=\delta _{2n}\}\subseteq K(\xi )$. Let us
assume the contrary. Then there exists $j\in \{1,\ldots ,2n\}$ such that the
equation $a_{j}=\delta _{j}$ does not belong to $K(\xi )$. Let $j\in
\{2i-1,2i\}$, where $i\in \{1,\ldots ,n\}$. Denote by $r_{0}$ the rule $(a_{2i-1}=0)\wedge
(a_{2i}=0)\rightarrow 0$ and by $r_{1}$ -- the rule $(a_{2i-1}=1)\wedge
(a_{2i}=1)\rightarrow 0$. It is clear that $r_{0},r_{1}\in S$ and at least
one of the equation systems $K(r_{0})\cup K(\xi )$ and $K(r_{1})\cup K(\xi )$
is consistent but this is impossible. Therefore $\{a_{1}=\delta _{1},\ldots
,a_{2n}=\delta _{2n}\}\subseteq K(\xi )$. From this relation and the inclusion $%
K(\xi )\subseteq K(S,\bar{\delta})$ it follows that in $\Gamma $ there are
at least $2^{n}$ pairwise different complete paths. Therefore $L(\Gamma
)\geq 2^{n}$ and $L_{SR}(S)\geq 2^{n}$.
\end{proof}

\begin{lemma}
\label{L10} Let $n,d\in \omega \setminus \{0\}$ and $d\geq 2$. Then there
exists a reduced decision rule system $S$ such that $n(S)=2n+d$, $k(S)=1$, $d(S)=d$,
$\left\vert S\right\vert =n+1$, and $L_{ESR}(S)\geq 2^{n}$.
\end{lemma}

\begin{proof}
Let us consider a system of decision rules $S=S_{1}\cup S_{2}$, where $%
S_{1}=\{(a_{2i-1}=0)\wedge (a_{2i}=0)\rightarrow 0:i=1,\ldots ,n\}$ and $%
S_{2}=\{(a_{2n+1}=0)\wedge \cdots \wedge (a_{2n+d}=0)\rightarrow 0\}$. It is
clear that $S$ is reduced, $n(S)=2n+d$, $d(S)=d$, $k(S)=1$, and $\left\vert S\right\vert =n+1$%
. Denote by $\Delta $ the set of tuples $(\delta _{1},\ldots ,\delta
_{2n+d})\in \{0,\ast \}^{2n+d}$ such that $\{\delta _{2i-1},\delta
_{2i}\}=\{0,\ast \}$ for $i=1,\ldots ,n$ and $\delta _{2n+1}=\cdots =\delta
_{2n+d}=\ast $. It is clear that $\left\vert \Delta \right\vert =2^{n}$ and,
for any $\bar{\delta}\in \Delta $, there is no a rule from $S$ that is
realizable for the tuple $\bar{\delta}$. Let $\Gamma $ be a decision tree
over $S$, which solves the problem $ESR(S)$ and for which $L(\Gamma
)=L_{ESR}(S)$. Let $\bar{\delta},\bar{\sigma}\in \Delta $ and $\bar{\delta}%
\neq \bar{\sigma}$. It is clear that there exist complete paths $\xi ,\tau $
in $\Gamma $ such that $K(\xi )\subseteq K(S,\bar{\delta})$ and $K(\tau
)\subseteq K(S,\bar{\sigma})$. Let us show that $\xi \neq \tau $. Let us
assume the contrary: $\xi =\tau $. It is clear that the terminal node of the
path $\xi $ is labeled with the empty set. Since $\bar{\delta}\neq \bar{%
\sigma}$, there exists $i\in \{1,\ldots ,n\}$ such that $(2i-1)$th and $(2i)$%
th digits of the tuples $\bar{\delta}$ and $\bar{\sigma}$ are different.
Therefore the attributes $a_{2i-1}$ and $a_{2i}$ are not attached to any node
of the path $\xi $. Hence the system of equations $K(r)\cup K(\xi )$ is
consistent, where $r$ is the rule $(a_{2i-1}=0)\wedge (a_{2i}=0)\rightarrow
0 $ from $S$, but this is impossible. Therefore $\xi \neq \tau $. Thus, in
the tree $\Gamma $, there are at least $2^{n}$ pairwise different complete
paths. As a result, we have $L(\Gamma )\geq 2^{n}$ and $L_{ESR}(S)\geq 2^{n}$%
.
\end{proof}

\begin{lemma}
\label{L11} (a) Let $n,d\in \omega \setminus \{0\}$. Then there exists a
reduced decision rule system $S$ such that $n(S)=n+d$, $k(S)=1$, $d(S)=d$, $%
\left\vert S\right\vert =n+1$, and $T_{EAD}(S)\geq 2^{n+1}$.

(b) Let $n,d,k\in \omega \setminus \{0\}$ and $k\geq 2$. Then there exists a reduced
decision rule system $S$ such that $n(S)=n+d$, $k(S)=k$, $d(S)=d$, $%
\left\vert S\right\vert =2n+k-1$, and $T_{AD}(S)\geq 2^{n}$.
\end{lemma}

\begin{proof}
(a) Let us consider a system of decision rules $S=\{(a_{1}=0)\rightarrow
1,(a_{2}=0)\rightarrow 2,\ldots ,(a_{n}=0)\rightarrow n,(a_{n+1}=0)\wedge
\cdots \wedge (a_{n+d}=0)\rightarrow n+1\}$. It is clear that $S$ is reduced, $n(S)=n+d$, $%
k(S)=1$, $d(S)=d$, $\left\vert S\right\vert =n+1$, and the problem $EAD(S)$
has $2^{n+1}$ pairwise different solutions. Therefore any decision tree
solving the problem $EAD(S)$ has at least $2^{n+1}$ terminal nodes that are
labeled with pairwise different sets of decision rules. Hence $%
T_{EAD}(S)\geq 2^{n+1}$.

(b) Let us consider a decision rule system $S=S_{1}\cup S_{2}\cup S_{3}$,
where $S_{1}=\{(a_{i}=0)\rightarrow 2i-1,(a_{i}=1)\rightarrow 2i:i=1,\ldots
,n\}$, $S_{2}=\{(a_{n+1}=0)\wedge \cdots \wedge (a_{n+d}=0)\rightarrow 0\}$,
and $S_{3}=\{(a_{1}=2)\rightarrow 0,\ldots ,(a_{1}=k-1)\rightarrow 0\}$. If $%
k=2$, then $S_{3}=0$. It is clear that $S$ is reduced, $n(S)=n+d$, $k(S)=k$, $d(S)=d$, and $%
\left\vert S\right\vert =2n+k-1$. It is also clear that the problem $AD(S)$
has at least $2^{n}$ pairwise different solutions. Therefore any decision
tree solving the problem $AD(S)$ has at least $2^{n}$ terminal nodes that
are labeled with pairwise different sets of decision rules. Hence $%
T_{AD}(S)\geq 2^{n}$.
\end{proof}

Let $k,d\in \omega \setminus \{0\}$. Denote $\Sigma (k,d)=\{S\in \Sigma :k(S)=k,d(S)=d\}$. We will consider only sequential algorithms for the construction of decision trees and will evaluate their time complexity depending on the size of decision rule systems on the algorithm input.

\begin{lemma}
\label{L12} Let $d\in \omega \setminus \{0\}$ and $C\in \{SR,AD,AR\}$. Then
there exists a polynomial time algorithm that, for a given decision rule
system $S\in \Sigma (1,d)$, constructs a decision tree solving the problem $%
C(S)$.
\end{lemma}

\begin{proof}
Let $S\in \Sigma (1,d)$, $A(S)=\{a_{i_{1}},\ldots ,a_{i_{m}}\}$, $%
i_{1}<\cdots <i_{m}$, and $V(S)=\{(\delta _{1},\ldots ,\delta _{m})\}$. We
construct a decision tree $\Gamma $ that consists of one complete path $\xi
=v_{1},d_{1},\ldots ,v_{m},d_{m},v_{m+1}$ such that, for $j=1,\ldots ,m$,
the node $v_{j}$ is labeled with the attribute $a_{i_{j}}$, the edge $d_{j}$
is labeled with the number $\delta _{j}$, and the node $v_{m+1}$ is labeled
with the set of decision rules $S$. It is clear that $\Gamma $ solves the
problem $C(S)$ and the considered algorithm has polynomial time complexity.
\end{proof}

\begin{lemma}
\label{L13} Let $k\in \omega \setminus \{0\}$ and $C\in \{SR,ESR\}$. Then
there exists a polynomial time algorithm that, for a given decision rule
system $S\in \Sigma (k,1)$, constructs a decision tree solving the problem $%
C(S)$.
\end{lemma}

\begin{proof}
Let $S\in \Sigma (k,1)$ and $S$ contain a rule $r$ of the length $0$. Construct a decision tree $\Gamma _0$ that contains only one node labeled with the set $\{r\}$. It is clear that $\Gamma _0$ solves the problem $C(S)$ and the considered
algorithm has polynomial time complexity.

Let us assume now that $S$ does not contain rules of the length $0$ and
 $A(S)=\{a_{i_{1}},\ldots ,a_{i_{n}}\}$. Let $%
j\in \{1,\ldots ,n\}$ and $\delta \in V_{S}(a_{i_{j}})$. It is clear that in
the system $S$ there exists a decision rule for which the left-hand side is
equal to $a_{i_{j}}=\delta $. Denote this rule by $r(j,\delta )$.

Let us construct a decision tree $\Gamma _{1}$. Let $V_{S}(a_{i_{1}})=\{%
\delta _{1},\ldots ,\delta _{m}\}$. The decision tree $\Gamma _{1}$ consists
of the nodes $v_{0},v_{1},\ldots ,v_{m}$ and the edges $d_{1},\ldots ,d_{m}$%
. The node $v_{0}$ is labeled with the attribute $a_{i_{1}}$, and the node $%
v_{j}$ is labeled with the set $\{r(1,\delta _{j})\}$, $j=1,\ldots ,m$. For $%
j=1,\ldots ,m$, the edge $d_{j}$ is labeled with the number $\delta _{j}$,
the edge $d_{j}$ leaves the node $v_{0}$ and enters the node $v_{j}$. It is
clear that $\Gamma _{1}$ solves the problem $SR(S)$ and the considered
algorithm has polynomial time complexity.

Let us construct a decision tree $\Gamma _{2}$. The tree $\Gamma _{2}$
contains a complete path $\xi =v_{1},d_{1},\ldots ,$ $v_{n},d_{n},v_{n+1}$.
For $j=1,\ldots ,n$, the node $v_{j}$ is labeled with the attribute $%
a_{i_{j}}$ and the edge $d_{j}$ is labeled with the symbol $\ast $. The node
$v_{n+1}$ is labeled with the empty set. Let $j\in \{1,\ldots ,n\}$ and $%
V_{S}(a_{i_{j}})=\{\delta _{1},\ldots ,\delta _{m}\}$. Besides the edge $%
d_{j}$, also the edges $d_{j1},\ldots ,d_{jm}$ leave the node $v_{j}$. These
edges are labeled with the numbers $\delta _{1},\ldots ,\delta _{m}$,
respectively. The edges $d_{j1},\ldots ,d_{jm}$ enter the nodes $%
v_{j1},\ldots ,v_{jm}$ that are labeled with the sets $\{r(j,\delta
_{1})\},\ldots ,\{r(j,\delta _{m})\}$, respectively. The tree $\Gamma $ does
not contain any other nodes and edges. It is clear that $\Gamma _{2}$ solves
the problem $ESR(S)$ and the considered algorithm has polynomial time
complexity.
\end{proof}

\section{Construction of Decision Trees \label{S4}}

As it was mentioned above, we consider only sequential algorithms for the construction of decision trees and evaluate their time complexity depending on the size of decision rule systems on the algorithm input. We  assume that, during each unit of time, the algorithm can add to the constructing decision tree at most one node.

Let $C\in \{SR,ESR,AD,EAD,AR,EAR\}$. Denote by $P_{DT}(C)$ the set of pairs $%
(k,d)\in (\omega \setminus \{0\})^{2}$ satisfying the following condition:
there exists a polynomial time algorithm that, for an arbitrary decision
rule system $S\in \Sigma (k,d)$, constructs a decision tree solving the
problem $C(S)$.

\begin{theorem}
\label{T1} (a) $P_{DT}(ESR)=\{(k,1):k\in \omega \setminus \{0\}\}$, (b) $%
P_{DT}(SR)=\{(k,1),(1,d):k,d\in \omega \setminus \{0\}\}$, (c) $%
P_{DT}(EAD)=P_{DT}(EAR)=\emptyset $, and (d) $P_{DT}(AD)=P_{DT}(AR)=\{(1,d):d\in
\omega \setminus \{0\}\}$.
\end{theorem}

\begin{proof}
Let $C\in \{SR,ESR,AD,EAD,AR,EAR\}$,  $(k,d)\in (\omega \setminus \{0\})^{2}$ and let us assume that there exists a polynomial time algorithm, which, for a given decision rule system $S\in \Sigma (k,d)$, constructs a decision tree solving the
problem $C(S)$. Then there exists a polynomial $\pi$ such that, for any $S\in \Sigma (k,d)$, $L_C(S) \le \pi (size(S))$. If $S$ is reduced, then, by (\ref{E13}), $size(S) \le10\left\vert S\right\vert
d(S)(\log _{2}(d(S)\left\vert S\right\vert )+1)$. We know that, for systems $S\in \Sigma (k,d)$, $d(S) \le d$. Therefore there exists a polynomial $\varrho$ such that $L_C(S) \le \varrho (\left\vert S\right\vert)$ for any reduced system $S\in \Sigma (k,d)$.

(a) Using
Lemma \ref{L13}, we obtain that $\{(k,1):k\in \omega \setminus
\{0\}\}\subseteq P_{DT}(ESR)$. Let $k,d\in \omega \setminus \{0\}$ and $d>1$%
. We now show that $(k,d)\notin P_{DT}(ESR)$. Let us assume the contrary: $%
(k,d)\in P_{DT}(ESR)$. Then there exists a polynomial $\varrho$ such that $L_{ESR}(S) \le \varrho (\left\vert S\right\vert)$ for any reduced system $S\in \Sigma (k,d)$.
Let $k=1$. Using Lemma \ref{L10}, we obtain that, for any $n\in \omega \setminus \{0\}$,  there exists a reduced decision rule system $S\in \Sigma (k,d)$ such that
$\left\vert S\right\vert =n+1$ and $L_{ESR}(S)\geq 2^{n}$. Therefore $2^{n} \le \varrho (n+1)$ for any $n\in \omega \setminus \{0\}$ but this is impossible.
Let $k>1$. Using Lemma \ref{L9}, we obtain that, for any $n\in \omega \setminus \{0\}$,  there exists a reduced decision rule system $S\in \Sigma (k,d)$ such that
$\left\vert S\right\vert =2n+k-1$ and $L_{SR}(S)\geq 2^{n}$. By Lemma \ref{L8}, $L_{ESR}(S)\geq 2^{n}$.
Therefore, $2^{n} \le \varrho (2n+k-1)$ for any $n\in \omega \setminus \{0\}$ but this is impossible since $k$ is a constant for the considered systems of decision rules.
Hence $(k,d)\notin P_{DT}(ESR)$ if $d>1$.
Thus, $P_{DT}(ESR)=\{(k,1):k\in \omega \setminus \{0\}\}$.

(b) Using
Lemmas \ref{L12} and \ref{L13}, we obtain that $\{(k,1),(1,d):k,d\in \omega \setminus
\{0\}\}\subseteq P_{DT}(SR)$. Let $k,d\in \omega \setminus \{0,1\}$. We now show that $(k,d)\notin P_{DT}(SR)$. Let us assume the contrary: $%
(k,d)\in P_{DT}(SR)$. Then there exists a polynomial $\varrho$ such that $L_{SR}(S) \le \varrho (\left\vert S\right\vert)$ for any reduced system $S\in \Sigma (k,d)$.
Using Lemma \ref{L9}, we obtain that, for any $n\in \omega \setminus \{0\}$,  there exists a reduced decision rule system $S\in \Sigma (k,d)$ such that
$\left\vert S\right\vert =2n+k-1$ and $L_{SR}(S)\geq 2^{n}$.
Therefore, $2^{n} \le \varrho (2n+k-1)$ for any $n\in \omega \setminus \{0\}$ but this is impossible since $k$ is a constant for the considered systems of decision rules.
Hence $(k,d)\notin P_{DT}(SR)$ if $k>1$ and $d>1$.
Thus,  $P_{DT}(SR)=\{(k,1),(1,d):k,d\in \omega \setminus \{0\}\}$.

(c) Let $k,d\in \omega \setminus \{0\}$. We now show that $(k,d)\notin P_{DT}(EAD)$. Let us assume the contrary: $%
(k,d)\in P_{DT}(EAD)$. Then there exists a polynomial $\varrho$ such that $L_{EAD}(S) \le \varrho (\left\vert S\right\vert)$ for any reduced system $S\in \Sigma (k,d)$.
Let $k=1$. Using Lemma \ref{L11}, we obtain that, for any $n\in \omega \setminus \{0\}$,  there exists a reduced decision rule system $S\in \Sigma (k,d)$ such that
$\left\vert S\right\vert =n+1$ and $T_{EAD}(S)\geq 2^{n+1}$. It is clear that $L_{EAD}(S)\geq T_{EAD}(S)$ and $L_{EAD}(S)\geq 2^{n+1}$.
Therefore, $2^{n+1} \le \varrho (n+1)$ for any $n\in \omega \setminus \{0\}$ but this is impossible.
Let $k>1$. Using Lemma \ref{L11}, we obtain that, for any $n\in \omega \setminus \{0\}$,  there exists a reduced decision rule system $S\in \Sigma (k,d)$ such that
$\left\vert S\right\vert =2n+k-1$ and $T_{AD}(S)\geq 2^{n}$. By Lemma \ref{L8}, $T_{EAD}(S)\geq T_{AD}(S)$. It is clear that $L_{EAD}(S)\geq T_{EAD}(S)$ and $L_{EAD}(S)\geq 2^{n}$.
Therefore, $2^{n} \le \varrho (2n+k-1)$ for any $n\in \omega \setminus \{0\}$ but this is impossible since $k$ is a constant for the considered systems of decision rules.
Hence $(k,d)\notin P_{DT}(EAD)$.
Thus, $P_{DT}(EAD)=\emptyset$.

Let $k,d\in \omega \setminus \{0\}$. We now show that $(k,d)\notin P_{DT}(EAR)$. Let us assume the contrary: $%
(k,d)\in P_{DT}(EAR)$. Then there exists a polynomial $\varrho$ such that $L_{EAR}(S) \le \varrho (\left\vert S\right\vert)$ for any reduced system $S\in \Sigma (k,d)$.
Let $k=1$. Using Lemma \ref{L11}, we obtain that, for any $n\in \omega \setminus \{0\}$,  there exists a reduced decision rule system $S\in \Sigma (k,d)$ such that
$\left\vert S\right\vert =n+1$ and $T_{EAD}(S)\geq 2^{n+1}$.  It is clear that $L_{EAD}(S)\geq T_{EAD}(S)$.  By Lemma \ref{L8}, $L_{EAR}(S)\geq L_{EAD}(S)$ and $L_{EAR}(S)\geq 2^{n+1}$.
Therefore, $2^{n+1} \le \varrho (n+1)$ for any $n\in \omega \setminus \{0\}$ but this is impossible.
Let $k>1$. Using Lemma \ref{L11}, we obtain that, for any $n\in \omega \setminus \{0\}$,  there exists a reduced decision rule system $S\in \Sigma (k,d)$ such that
$\left\vert S\right\vert =2n+k-1$ and $T_{AD}(S)\geq 2^{n}$. By Lemma \ref{L8}, $T_{EAR}(S)\geq T_{AD}(S)$. It is clear that $L_{EAR}(S)\geq T_{EAR}(S)$ and $L_{EAR}(S)\geq 2^{n}$.
Therefore, $2^{n} \le \varrho (2n+k-1)$ for any $n\in \omega \setminus \{0\}$ but this is impossible since $k$ is a constant for the considered systems of decision rules.
Hence $(k,d)\notin P_{DT}(EAR)$.
Thus, $P_{DT}(EAR)=\emptyset$.

(d) Using
Lemma \ref{L12}, we obtain that $\{(1,d):d\in \omega \setminus
\{0\}\}\subseteq P_{DT}(AD)$ and $\{(1,d):d\in \omega \setminus
\{0\}\}\subseteq P_{DT}(AR)$ .

Let $k,d\in \omega \setminus \{0\}$ and $k \ge 2$. We now show that $(k,d)\notin P_{DT}(AD)$. Let us assume the contrary: $%
(k,d)\in P_{DT}(AD)$. Then there exists a polynomial $\varrho$ such that $L_{AD}(S) \le \varrho (\left\vert S\right\vert)$ for any reduced system $S\in \Sigma (k,d)$.
Using Lemma \ref{L11}, we obtain that, for any $n\in \omega \setminus \{0\}$,  there exists a reduced decision rule system $S\in \Sigma (k,d)$ such that
$\left\vert S\right\vert =2n+k-1$ and $T_{AD}(S)\geq 2^{n}$.  It is clear that $L_{AD}(S)\geq T_{AD}(S)$ and $L_{AD}(S)\geq 2^{n}$.
Therefore, $2^{n} \le \varrho (2n+k-1)$ for any $n\in \omega \setminus \{0\}$ but this is impossible since $k$ is a constant for the considered systems of decision rules.
Hence $(k,d)\notin P_{DT}(AD)$.
Thus, $P_{DT}(AD)=\{(1,d):d\in
\omega \setminus \{0\}\}$.

Let $k,d\in \omega \setminus \{0\}$ and $k \ge 2$. We now show that $(k,d)\notin P_{DT}(AR)$. Let us assume the contrary: $%
(k,d)\in P_{DT}(AR)$. Then there exists a polynomial $\varrho$ such that $L_{AR}(S) \le \varrho (\left\vert S\right\vert)$ for any reduced system $S\in \Sigma (k,d)$.
Using Lemma \ref{L11}, we obtain that, for any $n\in \omega \setminus \{0\}$,  there exists a reduced decision rule system $S\in \Sigma (k,d)$ such that
$\left\vert S\right\vert =2n+k-1$ and $T_{AD}(S)\geq 2^{n}$.  It is clear that $L_{AD}(S)\geq T_{AD}(S)$. By Lemma \ref{L8}, $L_{AR}(S)\geq L_{AD}(S)$ and $L_{AR}(S)\geq 2^{n}$.
Therefore, $2^{n} \le \varrho (2n+k-1)$ for any $n\in \omega \setminus \{0\}$ but this is impossible since $k$ is a constant for the considered systems of decision rules.
Hence $(k,d)\notin P_{DT}(AR)$.
Thus, $P_{DT}(AR)=\{(1,d):d\in
\omega \setminus \{0\}\}$.
\end{proof}

\section{Construction of Acyclic Decision Graphs \label{S5}}

Besides decision trees, we will also consider \emph{acyclic decision graphs}
that are defined similarly to the decision trees, but instead of finite
directed trees with roots, finite directed graphs with roots that do not
have directed cycles are considered.  Note that any decision tree
is an acyclic decision graph. Let $C\in \{SR,ESR,AD,EAD,AR,EAR\}$. For a decision rule system  $S$, we denote by $L^{DG}_C(S)$ the minimum number of nodes in an acyclic decision graph, which solves the problem  $C(S)$. One can show that $L^{DG}_C(S) \ge T_C(S)$.

First, we describe a construction that will be used in this and in the next section.
Let $C\in \{SR,ESR,AD,EAD,AR,EAR\}$, $S \in \Sigma$, and $r\in S$ be a decision rule of the form $(a_{i_{1}}=\delta _{1})\wedge \cdots
\wedge (a_{i_{m}}=\delta _{m})\rightarrow \sigma $, where $m>0$.
For $j=1,\ldots ,m$, let $V_{S}^C(a_{i_{j}})=V_{S}(a_{i_{j}})$ if $C\in \{SR,AD,AR\}$ and $V_{S}^C(a_{i_{j}})=EV_{S}(a_{i_{j}})$ if $C\in \{ESR,EAD,EAR\}$.
We now
describe an acyclic decision graph $G_S^C(r)$. The graph $G_S^C(r)$ contains $m+2$ nodes
$v_{1},\ldots ,v_{m},v_{m+1},v_{m+2}$. The node $v_{1}$ is the root of $G_S^C(r)$%
. For $j=1,\ldots ,m$, the node $v_{j}$ is labeled with the attribute $%
a_{i_{j}}$, the node $v_{m+1}$ is labeled with the set $\{r\}$, and the node
$v_{m+2}$ is labeled with the empty set. For $j=1,\ldots ,m$, exactly $%
\left\vert V_{S}^C(a_{i_{j}})\right\vert $ edges leave the node $v_{j}$. These
edges are labeled with pairwise different elements from the set $%
V_{S}^C(a_{i_{j}})$. The edge labeled with the number $\delta _{j}$ enters the
node $v_{j+1}$. All other edges enter the node $v_{m+2}$. The graph $G_S^C(r)$
does not contain other nodes and edges.

We consider only sequential algorithms for the construction of acyclic decision graphs and evaluate their time complexity depending on the size of decision rule systems on the algorithm input. We  assume that, during each unit of time, the algorithm can add to the constructing acyclic decision graph at most one node.

Let $C\in \{SR,ESR,AD,EAD,AR,EAR\}$. Denote by $P_{DG}(C)$ the set of pairs $%
(k,d)\in (\omega \setminus \{0\})^{2}$ satisfying the following condition:
there exists a polynomial time algorithm that, for an arbitrary decision
rule system $S\in \Sigma (k,d)$, constructs an acyclic decision graph
solving the problem $C(S)$.

\begin{theorem}
\label{T2} (a) $P_{DG}(SR)=P_{DG}(ESR)=(\omega \setminus \{0\})^{2}$, (b) $%
P_{DG}(AD)=P_{DG}(AR)=\{(1,d):d\in \omega \setminus \{0\}\}$, and (c) $%
P_{DG}(EAD)=P_{DG}(EAR)=\emptyset $.
\end{theorem}

\begin{proof}
Let $C\in \{SR,ESR,AD,EAD,AR,EAR\}$ and  $(k,d)\in (\omega \setminus \{0\})^{2}$. Let us assume that there exists a polynomial time algorithm, which, for a given decision rule system $S\in \Sigma (k,d)$, constructs an acyclic decision graph solving the
problem $C(S)$. Then there exists a polynomial $\pi$ such that, for any $S\in \Sigma (k,d)$, $L^{DG}_C(S) \le \pi (size(S))$. If $S$ is reduced, then, by (\ref{E13}), $size(S) \le10\left\vert S\right\vert
d(S)(\log _{2}(d(S)\left\vert S\right\vert )+1)$. We know that, for systems $S\in \Sigma (k,d)$, $d(S) \le d$. Therefore there exists a polynomial $\varrho$ such that $T_C(S) \le L^{DG}_C(S) \le \varrho (\left\vert S\right\vert)$ for any reduced system $S\in \Sigma (k,d)$.

(a) Let $C \in \{SR,ESR\}$. We now show that $P_{DG}(C)=(\omega \setminus \{0\})^{2}$. Moreover, we
show that there exists a polynomial time algorithm, which, for a given
decision rule system $S\in \Sigma $, constructs an acyclic decision graph
solving the problem $C(S)$.

Let $S\in \Sigma $. If the system $S$ contains a decision rule $r$ of the
length $0$, then the acyclic decision graph consisting of one node that is
labeled with the set $\{r\}$ solves the problem $C(S)$. Let the system $S$ do not contain rules of the
length $0$ and $S=\{r_{1},\ldots ,r_{t}\}$. We connect the graphs $G_S^C(r_{1}),\ldots ,G_S^C(r_{t})$ (the
description of the graph $G_S^C(r)$ for a decision rule $r \in S$ can be found at the
beginning of Section \ref{S5}). To this end, for $j=1,\ldots ,t-1$,
replace the node of the graph $G_S^C(r_{j})$ labeled with the empty set with the
root of the graph $G_S^C(r_{j+1})$. Denote by $G_S^C$ the obtained graph. One can
show that the graph $G_S^C$ solves the problem $C(S)$. It is clear that the
considered algorithm has polynomial time complexity. Thus, $%
P_{DG}(C)=(\omega \setminus \{0\})^{2}$.

(b) Using
Lemma \ref{L12}, we obtain that $\{(1,d):d\in \omega \setminus
\{0\}\}\subseteq P_{DG}(AD)$ and $\{(1,d):d\in \omega \setminus
\{0\}\}\subseteq P_{DG}(AR)$ .

Let $k,d\in \omega \setminus \{0\}$ and $k \ge 2$. We now show that $(k,d)\notin P_{DG}(AD)$. Let us assume the contrary: $%
(k,d)\in P_{DG}(AD)$. Then there exists a polynomial $\varrho$ such that $T_{AD}(S) \le \varrho (\left\vert S\right\vert)$ for any reduced system $S\in \Sigma (k,d)$.
Using Lemma \ref{L11}, we obtain that, for any $n\in \omega \setminus \{0\}$,  there exists a reduced decision rule system $S\in \Sigma (k,d)$ such that
$\left\vert S\right\vert =2n+k-1$ and $T_{AD}(S)\geq 2^{n}$.
Therefore, $2^{n} \le \varrho (2n+k-1)$ for any $n\in \omega \setminus \{0\}$ but this is impossible since $k$ is a constant for the considered systems of decision rules.
Hence $(k,d)\notin P_{DG}(AD)$.
Thus, $P_{DG}(AD)=\{(1,d):d\in
\omega \setminus \{0\}\}$.

Let $k,d\in \omega \setminus \{0\}$ and $k \ge 2$. We now show that $(k,d)\notin P_{DG}(AR)$. Let us assume the contrary: $%
(k,d)\in P_{DG}(AR)$. Then there exists a polynomial $\varrho$ such that $T_{AR}(S) \le \varrho (\left\vert S\right\vert)$ for any reduced system $S\in \Sigma (k,d)$.
Using Lemma \ref{L11}, we obtain that, for any $n\in \omega \setminus \{0\}$,  there exists a reduced decision rule system $S\in \Sigma (k,d)$ such that
$\left\vert S\right\vert =2n+k-1$ and $T_{AD}(S)\geq 2^{n}$.   By Lemma \ref{L8}, $T_{AR}(S)\geq T_{AD}(S)$ and $T_{AR}(S)\geq 2^{n}$.
Therefore, $2^{n} \le \varrho (2n+k-1)$ for any $n\in \omega \setminus \{0\}$ but this is impossible since $k$ is a constant for the considered systems of decision rules.
Hence $(k,d)\notin P_{DT}(AR)$.
Thus, $P_{DG}(AR)=\{(1,d):d\in
\omega \setminus \{0\}\}$.

(c) Let $k,d\in \omega \setminus \{0\}$. We now show that $(k,d)\notin P_{DG}(EAD)$. Let us assume the contrary: $%
(k,d)\in P_{DG}(EAD)$. Then there exists a polynomial $\varrho$ such that $T_{EAD}(S) \le \varrho (\left\vert S\right\vert)$ for any reduced system $S\in \Sigma (k,d)$.
Let $k=1$. Using Lemma \ref{L11}, we obtain that, for any $n\in \omega \setminus \{0\}$,  there exists a reduced decision rule system $S\in \Sigma (k,d)$ such that
$\left\vert S\right\vert =n+1$ and $T_{EAD}(S)\geq 2^{n+1}$.
Therefore, $2^{n+1} \le \varrho (n+1)$ for any $n\in \omega \setminus \{0\}$ but this is impossible.
Let $k>1$. Using Lemma \ref{L11}, we obtain that, for any $n\in \omega \setminus \{0\}$,  there exists a reduced decision rule system $S\in \Sigma (k,d)$ such that
$\left\vert S\right\vert =2n+k-1$ and $T_{AD}(S)\geq 2^{n}$. By Lemma \ref{L8}, $T_{EAD}(S)\geq T_{AD}(S)$ and $T_{EAD}(S)\geq 2^{n}$.
Therefore, $2^{n} \le \varrho (2n+k-1)$ for any $n\in \omega \setminus \{0\}$ but this is impossible since $k$ is a constant for the considered systems of decision rules.
Hence $(k,d)\notin P_{DG}(EAD)$.
Thus, $P_{DG}(EAD)=\emptyset$.

Let $k,d\in \omega \setminus \{0\}$. We now show that $(k,d)\notin P_{DG}(EAR)$. Let us assume the contrary: $%
(k,d)\in P_{DG}(EAR)$. Then there exists a polynomial $\varrho$ such that $T_{EAR}(S) \le \varrho (\left\vert S\right\vert)$ for any reduced system $S\in \Sigma (k,d)$.
Let $k=1$. Using Lemma \ref{L11}, we obtain that, for any $n\in \omega \setminus \{0\}$,  there exists a reduced decision rule system $S\in \Sigma (k,d)$ such that
$\left\vert S\right\vert =n+1$ and $T_{EAD}(S)\geq 2^{n+1}$. By Lemma \ref{L8}, $T_{EAR}(S) \geq T_{EAD}(S)$  and $T_{EAR}(S)\geq 2^{n+1}$.
Therefore, $2^{n+1} \le \varrho (n+1)$ for any $n\in \omega \setminus \{0\}$ but this is impossible.
Let $k>1$. Using Lemma \ref{L11}, we obtain that, for any $n\in \omega \setminus \{0\}$,  there exists a reduced decision rule system $S\in \Sigma (k,d)$ such that
$\left\vert S\right\vert =2n+k-1$ and $T_{AD}(S)\geq 2^{n}$. By Lemma \ref{L8}, $T_{EAR}(S)\geq T_{AD}(S)$ and $T_{EAR}(S)\geq 2^{n}$.
Therefore, $2^{n} \le \varrho (2n+k-1)$ for any $n\in \omega \setminus \{0\}$ but this is impossible since $k$ is a constant for the considered systems of decision rules.
Hence $(k,d)\notin P_{DG}(EAR)$.
Thus, $P_{DG}(EAR)=\emptyset$.
\end{proof}

\section{Construction of Acyclic Decision Graphs with Writing \label{S6}}

Difficulties associated with a large number of pairwise different solutions to a problem can be circumvented by considering \emph{acyclic decision graphs with writing}, which, in addition to working nodes labeled with attributes, have \emph{writing } nodes and one terminal node. One of the nodes of the graph is distinguished as the root. Each writing node is labeled with a decision rule and has only one leaving edge. This edge is not labeled. The only terminal node is labeled with the letter $W$, denoting a set of decision rules $W$. This set can be changed during the work of the acyclic decision graph with writing. At the start of the work (when we are at the root of the graph), $W=\emptyset $. If during the work we come to a writing node that is labeled with a decision rule, we add this rule to the set $W$. When we reach the terminal node, the set $W$ formed by this moment is the result of the work of the considered acyclic decision graph with writing.

We consider only sequential algorithms for the construction of acyclic decision graphs with writing and evaluate their time complexity depending on the size of decision rule systems on the algorithm input. We  assume that, during each unit of time, the algorithm can add to the constructing acyclic decision graph with writing at most one node.

Let $C\in \{SR,ESR,AD,EAD,AR,EAR\}$. Denote by $P_{DGW}(C)$ the set of pairs
$(k,d)\in (\omega \setminus \{0\})^{2}$ satisfying the following condition:
there exists a polynomial time algorithm that, for an arbitrary decision
rule system $S\in \Sigma (k,d)$, constructs an acyclic decision graph with
writing solving the problem $C(S)$.

\begin{theorem}
\label{T3} $%
P_{DGW}(SR)=P_{DGW}(AD)=P_{DGW}(AR)=P_{DGW}(ESR)=P_{DGW}(EAD)=P_{DGW}(EAR)=(%
\omega \setminus \{0\})^{2}$.
\end{theorem}

\begin{proof}
Let $C\in \{SR,ESR,AD,EAD,AR,EAR\}$,  $S\in \Sigma$, and $r \in S$.
We now describe an acyclic decision graph with
writing $D_S^C(r)$. Let the length of the decision rule $r$ be equal to $0$.
Then $D_S^C(r)$ contains two nodes $v_{1}$ and $v_{2}$ and an edge that leaves
the node $v_{1}$ and enters the node $v_{2}$. The node $v_{1}$ is the root of $D_S^C(r)$. The node $v_{1}$ is labeled
with the decision rule $r$ and the node $v_{2}$ is labeled with the letter $%
W $. Let $r$ be a decision rule of the form $(a_{i_{1}}=\delta _{1})\wedge
\cdots \wedge (a_{i_{m}}=\delta _{m})\rightarrow \sigma $, where $m>0$. Then
the graph $D_S^C(r)$ is obtained from the graph $G_S^C(r)$ (this graph is defined at the beginning
of Section \ref{S5}) in the following way. Instead of the set $\{r\}$, we
label the node $v_{m+1}$ with the rule $r$. Instead of the empty set, we
label the node $v_{m+2}$ with the letter $W$. We add an edge leaving the node $%
v_{m+1}$ and entering the node $v_{m+2}$.

We now show that $P_{DGW}(C)=(\omega \setminus
\{0\})^{2}$. In fact, we show that there exists a polynomial time algorithm,
which, for a given decision rule system $S\in \Sigma $, constructs an
acyclic decision graph with writing $G$ that solves the problem $C(S)$.
Let $S=\{r_1, \ldots , r_t\}$.
First, we
construct graphs $D_S^C(r_{1}),\ldots ,D_S^C(r_{t})$. Then we connect these graphs.
To this end, for $j=1,\ldots ,t-1$, replace the node of the graph $D_S^C(r_{j})$
labeled with the letter $W$ with the root of the graph $D_S^C(r_{j+1})$. Denote
by $G$ the obtained acyclic decision graph with writing. One can show that
the graph $G$ solves the problem $C(S)$. It
is clear that the considered algorithm has polynomial time complexity.
\end{proof}

In connection with the difficulties arising in the construction of decision trees, it would be possible to move on to the study of acyclic decision graphs and acyclic decision graphs with writing. However, when constructing them, one should simultaneously try to make the depth as small as possible without allowing an excessive increase in the number of nodes. The possibilities of such bi-criteria optimization are the subject of a special study in the future. In the next section, we will consider another approach based on the simulation of the work of a decision tree on a given tuple of attribute values.

\section{Bounds and Algorithms Based on Node Covers\label{S7}}

In this section, we will continue the investigation of decision trees but instead of constructing the entire decision tree, we will restrict ourselves to the consideration of polynomial time algorithms that, for a given tuple of attribute values, describe the work of the decision tree on this tuple. To this end, we will study bounds on the depth of decision trees and algorithms for the description of the decision tree work based on node covers for the hypergraphs corresponding to the considered decision rule systems.

\subsection{Bounds\label{S7.1}}

In this section, we will study bounds on the minimum depth of decision trees solving problems. These bounds depend on the maximum length of the rule and parameters based on node covers for hypergraphs corresponding to the rule systems. In the next section, we will consider polynomial time algorithms for modeling the operation of a decision tree on a given tuple of attribute values that are based on the ideas proposed in this section.

Let $S$ be a decision rule system with $n(S)>0$. We denote by $S^{+}$ the subsystem of $S$
containing only rules of the length $d(S)$. Denote $\beta ^{+}(S)=\beta
(S^{+})$ where $\beta (S^{+})$ is the minimum cardinality of a node cover of
the hypergraph $G(S^{+})$. It is clear that $\beta (S^{+})\leq \beta (S)$.
Let $C\in \{EAR,EAD,ESR\}$. We denote by $E_{C}(S)$ the set of consistent
equation systems $\{a_{i_{1}}=\delta _{1},\ldots ,a_{i_{m}}=\delta _{m}\}$
such that $a_{i_{j}}\in A(S)$ and $\delta _{j}\in EV_{S}(a_{i_{j}})$ for $%
j=1,\ldots ,m$. Let $C\in \{AR,AD,SR\}$. We denote by $E_{C}(S)$ the set of
consistent equation systems $\{a_{i_{1}}=\delta _{1},\ldots
,a_{i_{m}}=\delta _{m}\}$ such that $a_{i_{j}}\in A(S)$ and $\delta _{j}\in
V_{S}(a_{i_{j}})$ for $j=1,\ldots ,m$. Denote $I_{EAD}(S)=I_{AD}(S)$ and $I_{ESR}(S)=I_{SR}(S)$.

Let $C\in \{AR,AD,SR,EAR\}$. Denote
\begin{eqnarray*}
\beta _{C}(S) &=&\max \{\beta (S_{\alpha }):\alpha \in E_{C}(S)\}, \\
\beta _{C}^{+}(S) &=&\max \{\beta ^{+}(S_{\alpha }):\alpha \in E_{C}(S)\}.
\end{eqnarray*}%
It is clear that $\beta _{C}^{+}(S)\leq \beta _{C}(S)$.

Let $C\in \{EAD,ESR\}$. Denote
\begin{eqnarray*}
\beta _{C}(S) &=&\max \{\beta (I_{C}(S_{\alpha })):\alpha \in E_{C}(S)\}, \\
\beta _{C}^{+}(S) &=&\max \{\beta ^{+}(I_{C}(S_{\alpha })):\alpha \in
E_{C}(S)\}.
\end{eqnarray*}%
It is clear that $\beta _{C}^{+}(S)\leq \beta _{C}(S)$.

\begin{lemma}
\label{L14}Let $S$ be a decision rule system with $n(S)>0$ and $C\in \{AR,EAR,EAD,ESR\}$.
Then $h_{C}(S)\geq \beta _{C}(S)\geq \beta _{C}^{+}(S)$.
\end{lemma}

\begin{proof}
Let $C\in \{AR,EAR\}$, $\alpha \in E_{C}(S)$, and $\beta _{C}(S)=\beta
(S_{\alpha })$. Using Lemma \ref{L4}, we obtain $h_{C}(S)\geq
h_{C}(S_{\alpha })$. By Lemma \ref{L5}, $h_{C}(S_{\alpha })\geq \beta
(S_{\alpha })=\beta _{C}(S)$. As we already mentioned, $\beta _{C}(S)\geq
\beta _{C}^{+}(S)$.

Let $C\in \{EAD,ESR\}$, $\alpha \in E_{C}(S)$, and $\beta _{C}(S)=\beta
(I_{C}(S_{\alpha }))$. Using Lemma \ref{L4}, we obtain $h_{C}(S)\geq
h_{C}(S_{\alpha })$. By Lemma \ref{L5}, $h_{C}(S_{\alpha })\geq \beta
(I_{C}(S_{\alpha }))=\beta _{C}(S)$. As we already mentioned, $\beta
_{C}(S)\geq \beta _{C}^{+}(S)$.
\end{proof}

\begin{lemma}
\label{L15} Let $S$ be a decision rule system with $n(S)>0$ and $C\in \{AR,EAR\}$. Then
\begin{equation*}
h_{C}(S)\leq d(S)\beta _{C}^{+}(S).
\end{equation*}
\end{lemma}

\begin{proof}
Let $V_{C}=V(S)$ if $C=AR$ and $V_{C}=EV(S)$ if $C=EAR$. We now describe the
work of a decision tree $\Gamma $ on a tuple from $V_{C}$. This work consists of rounds.

First round.
We construct a node cover $B_{1}$ of the hypergraph $G(S^{+})$ that has the
minimum cardinality, i.e., $\left\vert B_{1}\right\vert =\beta (S^{+})$. The
decision tree $\Gamma $ sequentially computes values of the attributes from $%
B_{1}$. As a result, we obtain a system $\alpha _{1}$ consisting of $%
\left\vert B_{1}\right\vert $ equations of the form $a_{i_{j}}=\delta _{j}$,
where $a_{i_{j}}\in B_{1}$ and $\delta _{j}$ is the computed value of the
attribute $a_{i_{j}}$. If $S_{\alpha _{1}}=\emptyset $ or all rules from $%
S_{\alpha _{1}}$ have the empty left-hand side, then the tree $\Gamma $
finishes its work. The result of this work is the set of decision rules $r$
from $S$ for which the system of equations $K(r)\cup \alpha _{1}$ is
consistent. Otherwise, we move on to the second round of the decision tree $%
\Gamma $ work.

Second round.
We construct a node cover $B_{2}$ of the hypergraph $G((S_{\alpha
_{1}})^{+}) $ that has the minimum cardinality, i.e., $\left\vert
B_{2}\right\vert =\beta ((S_{\alpha _{1}})^{+})$. The decision tree $\Gamma $
sequentially computes values of the attributes from $B_{2}$. As a result, we
obtain a system $\alpha _{2}$ consisting of $\left\vert B_{2}\right\vert $
equations. If $S_{\alpha _{1}\cup \alpha _{2}}=\emptyset $ or all rules from
$S_{\alpha _{1}\cup \alpha _{2}}$ have the empty left-hand side, then the
tree $\Gamma $ finishes its work. The result of this work is the set of
decision rules $r$ from $S$ for which the system of equations $K(r)\cup
\alpha _{1}\cup \alpha _{2}$ is consistent. Otherwise, we move on to the
third round of the decision tree $\Gamma $ work, etc.,  until we obtain empty system of rules or system in which all rules have empty left-hand side.

It is clear that $d(S)>d(S_{\alpha _{1}})>d(S_{\alpha _{1}\cup \alpha
_{2}})>\cdots $. Therefore, the number of rounds is at most $d(S)$. The
number of attributes values of which are computed by $\Gamma $ during each
round is at most $\beta _{C}^{+}(S)$. Therefore $h(\Gamma )\leq d(S)\beta
_{C}^{+}(S)$. It is easy to check that $\Gamma $ solves the problem $C(S)$.
Thus, $h_{C}(S)\leq d(S)\beta _{C}^{+}(S)$.
\end{proof}

\begin{lemma}
\label{L16} Let $S$ be a decision rule system with $n(S)>0$  and $C\in \{EAD,ESR\}$. Then
\begin{equation*}
h_{C}(S)\leq d(I_{C}(S))\beta _{C}^{+}(S).
\end{equation*}
\end{lemma}

\begin{proof}
Denote $V_{C}=EV(S)$. We now describe the work of a decision tree $\Gamma $
on a tuple from $V_{C}$. If all rules from $I_{C}(S)$ have the empty
left-hand side, then the tree $\Gamma $ finishes its work. The result of
this work is the set of decision rules $I_{C}(S)$. Otherwise, we move on to
the first round of the decision tree $\Gamma $ work.

We construct a node cover $B_{1}$ of the hypergraph $G(I_{C}(S)^{+})$ that
has the minimum cardinality, i.e., $\left\vert B_{1}\right\vert =\beta
(I_{C}(S)^{+})$. The decision tree $\Gamma $ sequentially computes values of
the attributes from $B_{1}$. As a result, we obtain a system $\alpha _{1}$
consisting of $\left\vert B_{1}\right\vert $ equations of the form $%
a_{i_{j}}=\delta _{j}$, where $a_{i_{j}}\in B_{1}$ and $\delta _{j}$ is the
computed value of the attribute $a_{i_{j}}$. If $I_{C}(S_{\alpha
_{1}})=\emptyset $ or all rules from $I_{C}(S_{\alpha _{1}})$ have the empty
left-hand side, then the tree $\Gamma $ finishes its work. The result of
this work is the set of decision rules $r$ from $S$ for which the system of
equations $K(r)\cup \alpha _{1}$ is consistent and $A(r)\subseteq B_{1}$.
Otherwise, we move on to the second round of the decision tree $\Gamma $
work.

We construct a node cover $B_{2}$ of the hypergraph $G(I_{C}(S_{\alpha
_{1}})^{+})$ that has the minimum cardinality, i.e., $\left\vert
B_{2}\right\vert =\beta (I_{C}(S_{\alpha _{1}})^{+})$. The decision tree $%
\Gamma $ sequentially computes values of the attributes from $B_{2}$. As a
result, we obtain a system $\alpha _{2}$ consisting of $\left\vert
B_{2}\right\vert $ equations. If $I_{C}(S_{\alpha _{1}\cup \alpha
_{2}})=\emptyset $ or all rules from $I_{C}(S_{\alpha _{1}\cup \alpha _{2}})$
have the empty left-hand side, then the tree $\Gamma $ finishes its work.
The result of this work is the set of decision rules $r$ from $S$ for which
the system of equations $K(r)\cup \alpha _{1}\cup \alpha _{2}$ is consistent
and $A(r)\subseteq B_{1}\cup B_{2}$. Otherwise, we move on to the third
round of the decision tree $\Gamma $ work, etc.,  until we obtain empty system of rules or system in which all rules have empty left-hand side.

One can show that $d(I_{C}(S))>d(I_{C}(S_{\alpha _{1}}))>d(I_{C}(S_{\alpha
_{1}\cup \alpha _{2}}))>\cdots $. Therefore the number of rounds is at most
$d(I_{C}(S))$. The number of attributes values of which are computed by $%
\Gamma $ during each round is at most $\beta _{C}^{+}(S)$. Therefore $%
h(\Gamma )\leq d(I_C(S))\beta _{C}^{+}(S)$. One can show that $\Gamma $
solves the problem $C(S)$. Thus, $h_{C}(S)\leq d(I_C(S))\beta _{C}^{+}(S)$.
\end{proof}

\begin{theorem}
\label{T4} Let $S$ be a decision rule system with $n(S)>0$. Then

(a) $\max \{d(S),\beta _{AR}(S)\}\leq h_{AR}(S)\leq d(S)\beta _{AR}^{+}(S)$.

(b) $\max \{d(S),\beta _{EAR}(S)\}\leq h_{EAR}(S)\leq d(S)\beta
_{EAR}^{+}(S) $.

(c) $h_{AD}(S)\leq d(S)\beta _{AD}^{+}(S)$.

(d) $\max \{d(S^{\prime }),\beta _{EAD}(S^{\prime })\}\leq h_{EAD}(S)\leq
d(S^{\prime })\beta _{EAD}^{+}(S^{\prime })$, where $S^{\prime }=R_{AD}(S)$.

(e) $h_{SR}(S)\leq d(S)\beta _{SR}^{+}(S)$.

(f) $\max \{d(S^{\prime }),\beta _{ESR}(S^{\prime })\}\leq h_{ESR}(S)\leq
d(S^{\prime })\beta _{ESR}^{+}(S^{\prime })$, where $S^{\prime }=R_{SR}(S)$.
\end{theorem}

\begin{proof}
(a) The lower bounds on $h_{AR}(S)$ follow from Lemmas \ref{L6} and \ref{L14}%
. The upper bound on $h_{AR}(S)$ follows from Lemma \ref{L15}.

(b) The lower bounds on $h_{EAR}(S)$ follow from Lemmas \ref{L6} and \ref%
{L14}. The upper bound on $h_{EAR}(S)$ follows from Lemma \ref{L15}.

(c) From Lemma \ref{L15} it follows that $h_{AR}(S)\leq d(S)\beta _{AR}^{+}(S)$%
. The inequality $h_{AD}(S)\leq h_{AR}(S)$ follows from Lemma \ref{L3}. It is
clear that $\beta _{AD}^{+}(S)=\beta _{AR}^{+}(S)$. Therefore $h_{AD}(S)\leq
d(S)\beta _{AD}^{+}(S)$.

(d) From Lemma \ref{L7} it follows that $h_{EAD}(S)=h_{EAD}(S^{\prime })$.
The bounds
\begin{equation*}
\max \{d(S^{\prime }),\beta _{EAD}(S^{\prime })\}\leq h_{EAD}(S^{\prime })
\end{equation*}%
follow from Lemmas \ref{L6} and \ref{L14}. The bound $h_{EAD}(S^{\prime})\leq d(I_{EAD}(S^{\prime }))\beta _{EAD}^{+}(S^{\prime })$ follows from Lemma \ref%
{L16}. It is clear that $I_{EAD}(S^{\prime })=S^{\prime}$.

(e) From Lemma  \ref{L15} it follows that $h_{AR}(S)\leq d(S)\beta _{AR}^{+}(S)$%
.The inequality $h_{SR}(S)\leq h_{AR}(S)$ follows from Lemma \ref{L3}. It is
clear that $\beta _{SR}^{+}(S)=\beta _{AR}^{+}(S)$. Therefore $h_{SR}(S)\leq
d(S)\beta _{SR}^{+}(S)$.

(f) From Lemma \ref{L7} it follows that $h_{ESR}(S)=h_{ESR}(S^{\prime })$.
The bounds
\begin{equation*}
\max \{d(S^{\prime }),\beta _{ESR}(S^{\prime })\}\leq h_{ESR}(S^{\prime })
\end{equation*}%
follow from Lemmas \ref{L6} and \ref{L14}. The bound $h_{ESR}(S^{\prime
})\leq d(I_{ESR}(S^{\prime }))\beta _{ESR}^{+}(S^{\prime })$ follows from Lemma \ref%
{L16}. It is clear that $I_{ESR}(S^{\prime })=S^{\prime}$.
\end{proof}

\begin{remark}
\label{R3} Note that the results for $h_{EAR}(S)$ and $h_{ESR}(S)$ mentioned in the
theorem were published in \cite{Moshkov01} without proofs.
\end{remark}

\subsection{Algorithms\label{S7.2}}

Algorithms described in the proofs of Lemmas \ref{L15} and \ref{L16} cannot be used in practice since they require construction of a node cover  of a hypergraph with minimum cardinality, which is an NP-hard problem. In this section, we consider a polynomial time algorithm for the construction of a node cover and modify algorithms described in the proofs of Lemmas \ref{L15} and \ref{L16}.

\subsubsection{Algorithm $\mathcal{A}_{cover}$\label{S7.2.1}}

Let $S$ be a decision rule system with $n(S)>0$ and $S^{+}$ be its subsystem consisting
of all rules from $S$ of the length $d(S)$. We now describe a polynomial
time algorithm $\mathcal{A}_{cover}$ for the construction of a node cover $B$
for the hypergraph $G(S^{+})$ such that $\left\vert B\right\vert \leq \beta
(S^{+})d(S)$.

\medskip \noindent
\emph{Algorithm} $\mathcal{A}_{cover}$
\medskip

\noindent
Set $B=\emptyset $.
We choose in $S^{+}$ an arbitrary rule $r_{1}$ and add all attributes from $%
A(r_{1})$ to $B$. We remove from $S^{+}$ all rules $r$ such that $%
A(r_{1})\cap A(r)\neq \emptyset $. Denote the obtained system by $S_{1}^{+}$%
. If $S_{1}^{+}=\emptyset $, then $B$ is a node cover of $G(S^{+})$. If $%
S_{1}^{+}\neq \emptyset $, then we choose in $S_{1}^{+}$ an arbitrary rule $%
r_{2} $ and add all attributes from $A(r_{2})$ to $B$. We remove from $%
S_{1}^{+}$ all rules $r$ such that $A(r_{2})\cap A(r)\neq \emptyset $.
Denote the obtained system by $S_{2}^{+}$. If $S_{2}^{+}=\emptyset $, then $B
$ is a node cover of $G(S^{+})$. If $S_{2}^{+}\neq \emptyset $, then we choose
in $S_{2}^{+}$ an arbitrary rule $r_{3}$, and so on until we construct a
node cover $B$. \medskip

Let $B=\bigcup_{i=1}^{t}A(r_{i})$. Since the sets $A(r_{1}),\ldots ,A(r_{t})$
are pairwise disjoint and the length of each of the rules $r_{1},\ldots
,r_{t}$ is equal to $d(S)$, we obtain $\left\vert B\right\vert =t\cdot d(S)$ and  $\beta
(S^{+})\geq t$. Therefore $\left\vert B\right\vert \leq \beta (S^{+})d(S)$.
One can show that the algorithm $\mathcal{A}_{cover}$ has polynomial time
complexity.

\subsubsection{Algorithm $\mathcal{A}_{tree}^{C}$, $C\in \{AR,EAR\}$\label{S7.2.2}}

Let $S$ be a decision rule system with $n(S)>0$, $C\in \{AR,EAR\}$, $V_{C}=V(S)$ if $C=AR$%
, and $V_{C}=EV(S)$ if $C=EAR$. We now describe a polynomial time algorithm $%
\mathcal{A}_{tree}^{C}$ that, for a given tuple of attribute values from the
set $V_{C}$, describes the work on this tuple of a decision tree $\Gamma $,
which solves the problem $C(S)$ and for which $h(\Gamma )\leq d(S)^{2}\beta
_{C}^{+}(S)$. This algorithm is a modification of the algorithm described in
the proof of Lemma \ref{L15}.

\medskip \noindent
\emph{Algorithm} $\mathcal{A}^C_{tree}$
\medskip

\noindent
The work of the decision tree $\Gamma$ consists of rounds.

First round. Using the algorithm $\mathcal{A}_{cover}$, we construct a node cover $B_{1}$
of the hypergraph $G(S^{+})$ with $\left\vert B_{1}\right\vert \leq
\beta (S^{+})d(S)$. The decision tree $\Gamma $ sequentially computes values
of the attributes from $B_{1}$. As a result, we obtain a system $\alpha _{1}$
consisting of $\left\vert B_{1}\right\vert $ equations of the form $%
a_{i_{j}}=\delta _{j}$, where $a_{i_{j}}\in B_{1}$ and $\delta _{j}$ is the
computed value of the attribute $a_{i_{j}}$. If $S_{\alpha _{1}}=\emptyset $
or all rules from $S_{\alpha _{1}}$ have the empty left-hand side, then the
tree $\Gamma $ finishes its work. The result of this work is the set of
decision rules $r$ from $S$ for which the system of equations $K(r)\cup
\alpha _{1}$ is consistent. Otherwise, we move on to the second round of the
decision tree $\Gamma $ work.

Second round. Using the algorithm $\mathcal{A}_{cover}$, we construct a node cover $B_{2}$
of the hypergraph $G((S_{\alpha _{1}})^{+})$ with $\left\vert
B_{2}\right\vert \leq \beta ((S_{\alpha _{1}})^{+})d(S_{\alpha _{1}})\leq
\beta ^{+}(S_{\alpha _{1}})d(S)$. The decision tree $\Gamma $ sequentially
computes values of the attributes from $B_{2}$. As a result, we obtain a
system $\alpha _{2}$ consisting of $\left\vert B_{2}\right\vert $ equations.
If $S_{\alpha _{1}\cup \alpha _{2}}=\emptyset $ or all rules from $S_{\alpha
_{1}\cup \alpha _{2}}$ have the empty left-hand side, then the tree $\Gamma $
finishes its work. The result of this work is the set of decision rules $r$
from $S$ for which the system of equations $K(r)\cup \alpha _{1}\cup \alpha
_{2}$ is consistent. Otherwise, we move on to the third round of the
decision tree $\Gamma $ work, etc.,  until we obtain empty system of rules  or system in which all rules have empty left-hand side. \medskip

It is clear that $d(S)>d(S_{\alpha _{1}})>d(S_{\alpha _{1}\cup \alpha
_{2}})>\cdots $. Therefore the number of rounds is at most $d(S)$. The
number of attributes values of which are computed by $\Gamma $ during each
round is at most $\beta _{C}^{+}(S)d(S)$. Therefore $h(\Gamma )\leq
d(S)^{2}\beta _{C}^{+}(S)$. It is easy to check that $\Gamma $ solves the
problem $C(S)$. One can show that the algorithm $\mathcal{A}_{tree}^{C}$ has
polynomial time complexity.

Let $S$ be a decision rule system with $n(S)>0$. For simplicity, we assume that $%
A(S)=\{a_{1},\ldots ,a_{n}\}$. We now show how the algorithms $\mathcal{A}%
_{tree}^{AR}$ and $\mathcal{A}_{tree}^{EAR}$ can be used for the description
of the work of the decision trees solving the problems $AR(S)$, $EAR(S)$, $%
AD(S)$, and $SR(S)$ on a tuple $\bar{\delta}=(\delta _{1},\ldots ,\delta
_{n})$ of values of attributes.

\emph{Problem }$AR(S)$. We apply the algorithm $\mathcal{A}_{tree}^{AR}$ to
the decision rule system $S$ and tuple $\bar{\delta}\in V(S)$. This
algorithm describes the work of a decision tree $\Gamma $, which solves the
problem $AR(S)$ and for which $h(\Gamma )\leq d(S)^{2}\beta _{AR}^{+}(S)$.
Using Lemma \ref{L6}, we obtain that $d(S)\leq h_{AR}(S)$. From Lemma \ref%
{L14} it follows that $\beta _{AR}^{+}(S)\leq h_{AR}(S)$. Therefore $%
h(\Gamma )\leq h_{AR}(S)^{3}$.

\emph{Problem }$EAR(S)$. We apply the algorithm $\mathcal{A}_{tree}^{EAR}$
to the decision rule system $S$ and tuple $\bar{\delta}\in EV(S)$. This
algorithm describes the work of a decision tree $\Gamma $, which solves the
problem $EAR(S)$ and for which $h(\Gamma )\leq d(S)^{2}\beta _{EAR}^{+}(S)$.
Using Lemma \ref{L6}, we obtain that $d(S)\leq h_{EAR}(S)$. From Lemma \ref%
{L14} it follows that $\beta _{EAR}^{+}(S)\leq h_{EAR}(S)$. Therefore $%
h(\Gamma )\leq h_{EAR}(S)^{3}$.

\emph{Problem }$AD(S)$. We apply the algorithm $\mathcal{A}_{tree}^{AR}$ to
the decision rule system $S$ and tuple $\bar{\delta}\in V(S)$. This
algorithm describes the work of a decision tree $\Gamma $, which solves the
problem $AR(S)$ and for which $h(\Gamma )\leq d(S)^{2}\beta _{AR}^{+}(S)$.
Using Lemma \ref{L2}, we obtain that $\Gamma $ solves the problem $AD(S)$.
It is clear that $\beta _{AR}^{+}(S)=\beta _{AD}^{+}(S)$. Therefore $%
h(\Gamma )\leq d(S)^{2}\beta _{AD}^{+}(S)$.

\emph{Problem }$SR(S)$. We apply the algorithm $\mathcal{A}_{tree}^{AR}$ to
the decision rule system $S$ and tuple $\bar{\delta}\in V(S)$. This
algorithm describes the work of a decision tree $\Gamma $, which solves the
problem $AR(S)$ and for which $h(\Gamma )\leq d(S)^{2}\beta _{AR}^{+}(S)$.
Using Lemma \ref{L2}, we obtain that $\Gamma $ solves the problem $SR(S)$.
It is clear that $\beta _{AR}^{+}(S)=\beta _{SR}^{+}(S)$. Therefore $%
h(\Gamma )\leq d(S)^{2}\beta _{SR}^{+}(S)$.

\subsubsection{Algorithm $\mathcal{B}_{tree}^{C}$, $C\in \{ESR,EAD\}$\label{S7.2.3}}

Let $S$ be a decision rule system with $n(S)>0$, $C\in \{ESR,EAD\}$, and $V_{C}=EV(S)$. We
now describe a polynomial time algorithm $\mathcal{B}_{tree}^{C}$ that, for
a given tuple of attribute values from the set $V_{C}$, describes the work
on this tuple of a decision tree $\Gamma $, which solves the problem $C(S)$
and for which $h(\Gamma )\leq d(I_{C}(S))^{2}\beta _{C}^{+}(S)$. This
algorithm is a modification of the algorithm described in the proof of Lemma %
\ref{L16}.

\medskip \noindent
\emph{Algorithm} $\mathcal{B}^C_{tree}$
\medskip

\noindent
If all rules from $I_{C}(S)$ have the empty left-hand side, then the tree $%
\Gamma $ finishes its work. The result of this work is the set of decision
rules $I_{C}(S)$. Otherwise, we move on to the first round of the decision
tree $\Gamma $ work.

Using the algorithm $\mathcal{A}_{cover}$, we construct a node cover $B_{1}$
of the hypergraph $G(I_{C}(S)^{+})$ with $\left\vert
B_{1}\right\vert \leq \beta (I_{C}(S)^{+})d(I_{C}(S))$. The decision tree $%
\Gamma $ sequentially computes values of the attributes from $B_{1}$. As a
result, we obtain a system $\alpha _{1}$ consisting of $\left\vert
B_{1}\right\vert $ equations of the form $a_{i_{j}}=\delta _{j}$, where $%
a_{i_{j}}\in B_{1}$ and $\delta _{j}$ is the computed value of the attribute
$a_{i_{j}}$. If $I_{C}(S_{\alpha _{1}})=\emptyset $ or all rules from $%
I_{C}(S_{\alpha _{1}})$ have the empty left-hand side, then the tree $\Gamma
$ finishes its work. The result of this work is the set of decision rules $r$
from $S$ for which the system of equations $K(r)\cup \alpha _{1}$ is
consistent and $A(r)\subseteq B_{1}$. Otherwise, we move on to the second
round of the decision tree $\Gamma $ work.

Using the algorithm $\mathcal{A}_{cover}$, we construct a node cover $B_{2}$
of the hypergraph $G(I_{C}(S_{\alpha _{1}})^{+})$ with $\left\vert
B_{2}\right\vert \leq \beta (I_{C}(S_{\alpha _{1}})^{+})d(I_{C}(S_{\alpha
_{1}}))\leq \beta (I_{C}(S_{\alpha _{1}})^{+})d(I_{C}(S))$. The decision
tree $\Gamma $ sequentially computes values of the attributes from $B_{2}$.
As a result, we obtain a system $\alpha _{2}$ consisting of $\left\vert
B_{2}\right\vert $ equations. If $I_{C}(S_{\alpha _{1}\cup \alpha
_{2}})=\emptyset $ or all rules from $I_{C}(S_{\alpha _{1}\cup \alpha _{2}})$
have the empty left-hand side, then the tree $\Gamma $ finishes its work.
The result of this work is the set of decision rules $r$ from $S$ for which
the system of equations $K(r)\cup \alpha _{1}\cup \alpha _{2}$ is consistent
and $A(r)\subseteq B_{1}\cup B_{2}$. Otherwise, we move on to the third
round of the decision tree $\Gamma $ work, etc.,  until we obtain empty system of rules or system in which all rules have empty left-hand side. \medskip

One can show that $d(I_{C}(S))>d(I_{C}(S_{\alpha _{1}}))>d(I_{C}(S_{\alpha
_{1}\cup \alpha _{2}}))>\cdots $. Therefore the number of rounds is at most
$d(I_{C}(S))$. The number of attributes values of which are computed by $%
\Gamma $ during each round is at most $\beta _{C}^{+}(S)d(I_{C}(S))$.
Therefore $h(\Gamma )\leq d(I_{C}(S))^{2}\beta _{C}^{+}(S)$. One can show that $\Gamma $ solves the problem $C(S)$. It is easy to check that the
algorithm $\mathcal{B}_{tree}^{C}$ has polynomial time complexity.

Let $S$ be a decision rule system with $n(S)>0$. For simplicity, we assume that $%
A(S)=\{a_{1},\ldots ,a_{n}\}$. We now show how the algorithms $\mathcal{B}%
_{tree}^{EAD}$ and $\mathcal{B}_{tree}^{ESR}$ can be used for the
description of the work of the decision trees solving the problems $EAD(S)$
and $ESR(S)$ on a tuple $\bar{\delta}=(\delta _{1},\ldots ,\delta _{n})$ of
values of attributes.

\emph{Problem }$EAD(S)$. Construct the rule system $S^{\prime }=R_{AD}(S)$.
We apply the algorithm $\mathcal{B}_{tree}^{EAD}$ to the decision rule
system $S^{\prime }$ and adapt it to the work with tuple $\bar{\delta}%
=(\delta _{1},\ldots ,\delta _{n})\in EV(S)$. From the description of the
algorithm $\mathcal{B}_{tree}^{EAD}$ it follows that it will not compute
values of the attributes from $A(S)\setminus A(S^{\prime })$. Let the algorithm $\mathcal{B}_{tree}^{EAR}$ should compute the value of
an attribute $a_{i}\in A(S^{\prime })$. If $\delta _{i}\in EV_{S^{\prime
}}(a_{i})$, then $\mathcal{B}_{tree}^{EAD}$ will work normally. If $\delta
_{i}\notin EV_{S^{\prime }}(a_{i})$, then $\mathcal{B}_{tree}^{EAD}$ will work
in the same way as in the case $\delta _{i}=\ast $. One can show that the
adapted algorithm $\mathcal{B}_{tree}^{EAD}$ describes the work of a
decision tree $\Gamma $, which solves the problem $EAD(S)$ and for which $%
h(\Gamma )\leq d(I_{EAD}(S^{\prime }))^{2}\beta _{EAD}^{+}(S^{\prime })$.
Using Lemma \ref{L6}, we obtain that $d(I_{EAD}(S^{\prime }))\leq d(S^{\prime
})\leq h_{EAD}(S^{\prime })$. From Lemma \ref{L14} it follows that $\beta
_{EAD}^{+}(S^{\prime })\leq h_{EAD}(S^{\prime })$. Therefore $h(\Gamma )\leq
h_{EAD}(S^{\prime })^{3}$. From Lemma \ref{L7} it follows that $%
h_{EAD}(S)=h_{EAD}(S^{\prime })$. Thus, $h(\Gamma )\leq h_{EAD}(S)^{3}$.

\emph{Problem }$ESR(S)$. Construct the rule system $S^{\prime }=R_{SR}(S)$.
We apply the algorithm $\mathcal{B}_{tree}^{ESR}$ to the decision rule
system $S^{\prime }$ and adapt it to the work with tuple $\bar{\delta}%
=(\delta _{1},\ldots ,\delta _{n})\in EV(S)$. From the description of the
algorithm $\mathcal{B}_{tree}^{ESR}$ it follows that it will not compute
values of the attributes from $A(S)\setminus A(S^{\prime })$. Let the algorithm $\mathcal{B}_{tree}^{ESR}$ should compute the value of
an attribute $a_{i}\in A(S^{\prime })$. If $\delta _{i}\in EV_{S^{\prime
}}(a_{i})$, then $\mathcal{B}_{tree}^{ESR}$ will work normally. If $\delta
_{i}\notin EV_{S^{\prime }}(a_{i})$, then $\mathcal{B}_{tree}^{ESR}$ will work
in the same way as in the case $\delta _{i}=\ast $. One can show that the
adapted algorithm $\mathcal{B}_{tree}^{ESR}$ describes the work of a
decision tree $\Gamma $, which solves the problem $ESR(S)$ and for which $%
h(\Gamma )\leq d(I_{ESR}(S^{\prime }))^{2}\beta _{ESR}^{+}(S^{\prime })$.
Using Lemma \ref{L6}, we obtain that $d(I_{ESR}(S^{\prime }))\leq d(S^{\prime
})\leq h_{ESR}(S^{\prime })$. From Lemma \ref{L14} it follows that $\beta
_{ESR}^{+}(S^{\prime })\leq h_{ESR}(S^{\prime })$. Therefore $h(\Gamma )\leq
h_{ESR}(S^{\prime })^{3}$. From Lemma \ref{L7} it follows that $%
h_{ESR}(S)=h_{ESR}(S^{\prime })$. Thus, $h(\Gamma )\leq h_{ESR}(S)^{3}$.

\begin{remark}
\label{R4}Note that the results for the problems $EAR(S)$ and $ESR(S)$ similar to
mentioned above were published in \cite{Moshkov01} without proofs.
\end{remark}

\section{Conclusion\label{S8}}
In this paper, we considered the problem of
constructing decision trees and acyclic decision graphs representing
decision trees for given rule systems, and discussed the possibility of
constructing not the entire decision tree, but the computation path in this
tree for the given input. The future work will be focused on the
dynamic programming and greedy algorithms for the construction
of decision trees for given decision rule systems.

\subsection*{Acknowledgements}

Research reported in this publication was
supported by King Abdullah University of Science and Technology (KAUST).

\bibliographystyle{spmpsci}
\bibliography{1C}

\end{document}